\documentclass{article}

\PassOptionsToPackage{numbers,sort&compress}{natbib}
\PassOptionsToPackage{table}{xcolor}
\PassOptionsToPackage{hyphens}{url}
\usepackage[preprint]{neurips_2026}

\usepackage[utf8]{inputenc}
\usepackage[T1]{fontenc}
\usepackage{amsmath}
\usepackage{amsfonts}
\usepackage{amssymb}
\usepackage{microtype}
\usepackage{hyperref}
\usepackage{xurl}
\usepackage{booktabs}
\usepackage[most]{tcolorbox}
\usepackage{ulem}
\usepackage{capt-of}
\usepackage{placeins}
\usepackage{wrapfig}
\usepackage{needspace}
\usepackage{dsfont}
\usepackage{xcolor}
\usepackage{array}
\usepackage{fvextra}
\DeclareUnicodeCharacter{2264}{\ensuremath{\le}}
\newcommand{\draftblue}[1]{\textcolor{blue}{#1}}
\newcommand{\tbd}[1]{\textcolor{orange}{#1}}
\definecolor{DarkGreen}{RGB}{0,90,50}

\usepackage{xspace}

\newcommand{\ie}{\textit{i.e.,}\xspace}
\newcommand{\eg}{\textit{e.g.,}\xspace}

\newcommand{\method}{\textsc{DisCorD}\xspace}
\newcommand{\methodd}{\textsc{DisCorD}}

\definecolor{darkblue}{rgb}{0, 0, 0.5}
\hypersetup{colorlinks=true, citecolor=darkblue, linkcolor=darkblue, urlcolor=darkblue}

\title{Distribution Corrected Offline Data Distillation for Large Language Models}
\author{Yumeng Zhang, Zhengbang Yang, Yevin Nikhel Goonatilake, Zhuangdi Zhu\\
George Mason University, Fairfax, VA, USA}

\usepackage{graphicx}

\begin{document}

\maketitle

\begin{abstract}
Distilling reasoning traces from strong large language models into smaller ones is a promising route to improve intelligence in resource-constrained settings. Existing approaches face a fundamental trade-off: offline distillation from teacher-generated traces provides high-quality, sample-efficient supervision but suffers from distributional drift: during training, the student model conditions on teacher-generated prefixes, whereas during inference the student autoregresses on self-generated prefixes, leading to compounding errors over long reasoning trajectories. Meanwhile, on-policy or self-distillation methods better match the student’s inference-time distribution, but require costly online sampling and often produce low-quality traces in early training.
We propose a principled offline reasoning distillation framework that preserves the efficiency and supervision quality of offline teacher-generated data while correcting teacher–student distribution drift. It adaptively emphasizes teacher supervision that is better aligned with the student's on-policy distribution.
Evaluations on mathematical reasoning benchmarks of GSM8K, MATH, MATH500, and harder held-out competition-style tasks, including AMC, AIME, and OlympiadBench, show that our method improves reasoning accuracy over prior offline distillation algorithms and yields more stable reasoning traces while preserving instruction-following capabilities.
Our work shows that lightweight, distribution-correction-aware training can substantially strengthen offline reasoning distillation without online rollouts.\footnote{Code is available at \url{https://anonymous.4open.science/r/Discord-E6BC}.}

%
%
\end{abstract}

\vspace{-0.1in}
\section{Introduction} \vspace{-0.1in}
Reasoning ability has emerged as a cornerstone of modern large language models (LLMs),
largely driven by scaled model architecture and advances in reinforcement learning (RL)-based post-training optimization. This capability is particularly important in scientific and engineering domains such as mathematics, robotics, and physics, where solving problems requires structured, multi-step reasoning instead of short-horizon token prediction.
%
Despite these advances, high-quality reasoning behavior is still primarily observed in large or proprietary models. Enabling smaller models to acquire similar reasoning ability is critical for practical deployment, especially in resource-constrained environments where efficient and reliable intelligence is required. The ability to transfer reasoning competence across models without significant performance degradation is crucial.

A common approach to reasoning distillation is to train a smaller student model on reasoning traces generated by a stronger teacher model. In this paradigm, referred to as \textit{offline learning} or \textit{behavior cloning}, training data are collected from an external model. In contrast, \textit{on-policy} methods generate data from the target model itself and update the model using policy-gradient RL techniques such as GRPO \cite{shao2024deepseekmathpushinglimitsmathematical} or token-level extensions \cite{liu2025tisdpo,yang2026tokenimportance,zeng2024tdpo,yoon-etal-2024-tlcr}.

Offline distillation offers several advantages. It is typically more sample-efficient, and teacher-generated trajectories provide higher-quality supervision, especially during early stages of training.
However, it also faces the inherent limitation of \textit{distribution drift}: the distribution of teacher-generated trajectories differs from the distribution induced by the student model during inference. As the student generates intermediate tokens that deviate from teacher trajectories, it enters support regions that are underrepresented in the training data. This distribution shift leads to compounding errors and unreliable behavior. Existing approaches alleviate this issue through refined distillation objectives, such as bidirectional divergence minimization \cite{jung2025todi}, but these methods do not directly resolve the distribution mismatch, where the optimization is still focused on a teacher-provided distribution.

We propose a principled method for LLM reasoning distillation that relies solely on offline, teacher-provided data while reducing the distribution gap to on-policy training. %
We first design a general reasoning-eliciting instruction that contextualizes a large teacher model to produce high-quality reasoning traces. These traces serve as supervision for training a smaller student model.
We then introduce a distribution-corrected objective that incorporates an advantage-based weighting mechanism, enabling the student to better align with trajectories it is likely to encounter during inference.
Our formulation provides an effective approach for reducing the teacher–student distributional drift and is compatible with most offline distillation frameworks.
We focus on mathematical reasoning as a representative domain. Empirical results across multiple models and math benchmarks demonstrate that our method consistently improves the reasoning performance of student models in both in-domain and out-of-domain settings, outperforming existing offline distillation baselines and narrowing the gap to that of online sampling methods without requiring online interaction.

\section{Preliminary} \vspace{-0.1in}
\paragraph{On-policy  Learning.}
On-policy methods, especially on-policy RL, have been widely adopted for LLM post-training in domains such as math reasoning and safety alignment.
The core idea behind this approach is to fine-tune a pretrained model $\pi_\theta $ to maximize accumulated rewards within a domain $\mathcal{Q} $ over self-generated trajectories:
\begin{equation}
    J_{\text{RL}}(\theta)
    = \mathbb{E}_{x \sim \mathcal{Q},\; y \sim \pi_\theta(\cdot \mid x)}
      \left[ \mathds{R}(y, x) \right],
    \label{eq:onpolicy_obj}
\end{equation}
where $\mathds{R}(x,y)$ is a reward function that scores a response $y$ (\textit{e.g.}, a reasoning trajectory).
Following the policy gradient principle~\cite{sutton1999policy}, we derive the following gradient update for Eq.~\eqref{eq:onpolicy_obj}:
\begin{small}
\begin{equation}
    \nabla_\theta J_{\text{RL}}(\theta)
    = \mathbb{E}_{x \sim \mathcal{Q},\; \underbrace{y \sim \pi_\theta(\cdot \mid x)}
     _{\text{On-policy sampling}}} \left[ \mathds{R}(x,y) \cdot
     \sum_{t=0}^{|y|} \log \pi_\theta(y_t \mid x,\, y_{:<t})  \right],
    \label{eq:onpolicy_grad}
\end{equation}\end{small}
Prior work has improved training stability by replacing the reward function $\mathds{R}(x,y)$ with a normalized  \textit{Advantage} function, tokenizing the optimization formulation, and introducing clipping and model update constraints~\cite{shao2024deepseekmathpushinglimitsmathematical,guo2025deepseekr1,liu2025tisdpo,yang2026tokenimportance,zeng2024tdpo,yoon-etal-2024-tlcr}.

A key advantage of this strategy over supervised fine-tuning (SFT) lies in its \textit{on-policy} learning: the data used to optimize the target model is sampled from itself instead of  from another behavior policy.
It mitigates distribution drift, where the model encounters underrepresented regions at inference time, and often benefits out-of-domain generalization.

However, online methods are costly in terms of sample collection and labeling, requiring a notably higher number of training samples than supervised fine-tuning, and each trajectory requires labeled rewards.
Additionally, for smaller LLMs, the sample quality is also limited by the model's own capability, which leads to rare high-return trajectories and slow asymptotic convergence.

\vspace{-0.1in}
\paragraph{Offline Knowledge Distillation.}
A parallel path towards improving pretrained LLM performance is knowledge distillation from another stronger model, such as an advanced teacher LLM $\pi_T$.
One straightforward method is to directly perform SFT on teacher-generated reasoning traces,  which trains the target (student) model $\pi_\theta$ to imitate teacher behavior by maximizing the log-likelihood of  \textit{\textbf{teacher-generated}} responses:
\begin{small}\begin{equation}
    J_{\text{SFT}}(\theta)
    = \mathbb{E}_{ x \sim \mathcal{Q},\; \underbrace{y \sim \pi_T(\cdot \mid x)}_{\text{off-line sampling}}}
      \left[ \sum_{t=0}^{|y|} \log \pi_\theta(y_t \mid x,\, y_{:<t}) \right].
    \label{eq:offline_obj}
\end{equation}\end{small}
Here the expectation is taken under the \textit{teacher} distribution $\pi_T$, with $y$  drawn from an offline dataset provided by the teacher model.
This \textit{behavior-cloning} style method is training- and sampling-efficient but is also prone to performance drop induced by distribution drift, since this offline objective does not directly optimize the target ($\pi_\theta$) model's performance when $y$ is sampled from $\pi_\theta$ during inference.

Subsequent  methods supervise the student model with more refined  teacher signals,  often by minimizing a distributional divergence between the teacher and student models, such as the KL-divergence:
\vspace{-0.1in}\begin{align}
 J_{\text{KD}}(\theta)
    =  \mathbb{E}_{x \sim \mathcal{Q},\; y \sim \pi_T(\cdot \mid x)}
      \left[  \mathcal{D}_\text{KL}(\pi_T(y|x) \vert \pi_\theta (y|x)) \right].
    \label{eq:offline_kd}
\end{align}
Works along this line include optimizing on a {reverse} KL divergence $\mathcal{D}_{\text{KL}}( \pi_\theta  \Vert \pi_T )$~\cite{cao2025llm} or a mixture of both directions \cite{jung2025todi}.
While these algorithms show empirical performance improvement over direct SFT,  they still inherit a fundamental limitation of offline data: the training objective provides a biased signal for the student's on-policy behavior.
At inference time,  the student model may generate prefixes that are underrepresented in the offline teacher data, leading to unreliable behavior \cite{agarwal2024onpolicy}.

\begin{figure}[t!]
    \centering
    \includegraphics[width=1\textwidth]{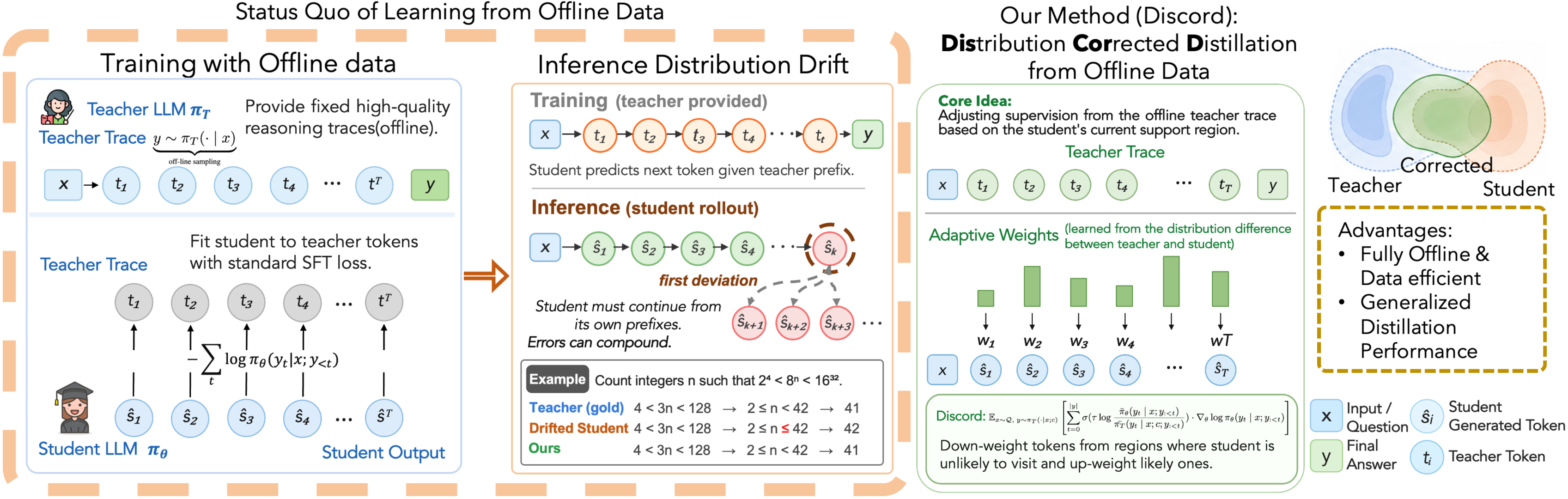}
    \vspace{-0.1in} \caption{\small{Overview of \method:
Standard offline distillation imitates fixed teacher traces under teacher-generated prefixes, but during inference the student conditions on its own prefixes, causing distribution drift and compounding errors (left). \method{} reduces this mismatch by reweighting offline teacher-token supervision according to the student's support, emphasizing teacher behaviors that are more reachable under the student's inference-time distribution (right).
}} \vspace{-0.2in}
    \label{fig:framework_overview}
\end{figure}

\vspace{-0.1in}
\section{Methodology: Distribution Corrected Distillation from Offline Data } \vspace{-0.1in}
In this paper, we aim to improve cross-model knowledge distillation for reasoning, where a teacher LLM with more advanced reasoning ability provides supervision for a student LLM.
To elicit strong teacher behavior, we first design a reasoning instruction as a conditional context $c$ when sampling traces from the teacher: $y \sim \pi_T(x;c)$, which forms an offline reasoning dataset for student distillation.

Our goal is to retain the data-efficiency benefits of offline distillation while reducing the distributional gap between teacher-generated traces and the student's on-policy behavior.
The core idea is to introduce a student-support-aware correction that reweights a training trajectory $\{x, y\sim\pi_T(x;c)\} $
with a density ratio $\frac{\bar{\pi}_\theta(y \mid x)}{\bar{\pi}_T(y \mid x;c)}$, where $\bar{\pi}$ denotes a stop-gradient operation.

\vspace{-0.1in}
\paragraph{Stabilized Transformation.}
However, directly applying such a density ratio  leads to high training instability, as the density ratio is unbounded  and accumulates high variance over long sequences. We address this in two steps.
First, to obtain a bounded and monotonic approximation, we apply a sigmoid transformation to the log density ratio:
$r(y \mid x) = \sigma\!\left(\tau \cdot \log \frac{\bar{\pi}_\theta(y \mid x)}{\bar{\pi}_T(y \mid x; c)}\right),$
with a  temperature coefficient $\tau$ set to 1 by default.
It monotonically preserves the ordering induced by the density ratio while mapping the correction into a bounded range of $(0,1)$.

Meanwhile, directly applying $r(y \mid x) $ as a sequence-level weight still introduces length bias, where longer trajectories can accumulate larger aggregate log-probability differences that inflate their correction weights.
To mitigate such length bias and variance, we apply token-level approximation, following a strategy that has also been adopted to stabilize RL training~\cite{yu2025dapo}.
Specifically, we tokenize the correction by applying it as gradient weighting $r_t$ for each training token $y_t$ in a trajectory $y=\{y_{t}\}_{t<|y|}$.
This leads to our final reweighting ratio: $r_t
    =
    \sigma\left(
    \tau \log
    \frac{
    \bar{\pi}_\theta(y_t \mid x; y_{:<t})
    }{
    \bar{\pi}_T(y_t \mid x; c; y_{:<t})
    }
    \right).$

This formulation can be readily applied to different offline distillation backbones.
We illustrate with SFT as the base objective:
\vspace{-0.1in}\begin{small}\begin{equation}
    \nabla_\theta J^{\text{\method}}(\theta)
 = \mathbb{E}_{x \sim \mathcal{Q},\; y \sim \pi_T(\cdot \mid x;c)}
      \left[\sum_{t=0}^{|y|}
        \sigma(\tau \log \frac{\bar{\pi}_\theta(y_t \mid x; y_{:<t})}{\bar{\pi_T}(y_t \mid x;c; y_{:<t})})
        \cdot \nabla_\theta \log \pi_\theta(y_t \mid x; y_{:<t})
      \right].
    \label{eq:grad_ideal}
\end{equation}\end{small}

We refer to the proposed framework as \method (\textbf{Dis}tribution-\textbf{Cor}rected \textbf{D}istillation), which modulates the influence of teacher-provided data supervision based on the student's current support.
It is also compatible with other offline distillation objectives beyond SFT, such as the KL-based formulation in Eq.~\eqref{eq:offline_kd} by replacing the gradient of SFT with that of KL  minimization (Section~\ref{sec:experiments}).
%
%
\vspace{-0.05in}
\paragraph{Implications of the Token-Level Correction.}
The correction weight at token $t $ takes the form using a default temperature of 1:
\begin{small}\begin{align*}
r_t|_{\tau=1} = \sigma\!\left(\log \frac{\bar{\pi}_\theta(y_t \mid x; y_{:<t})}{\bar{\pi}_T(y_t \mid x; c; y_{:<t})}\right) = \frac{\bar{\pi}_\theta(y_t \mid x; y_{:<t})}{\bar{\pi}_\theta(y_t \mid x; y_{:<t}) + \bar{\pi}_T(y_t \mid x; c; y_{:<t})}. \end{align*}\end{small}
It can be interpreted as the posterior probability that token $y_t $ was drawn from the student distribution $\pi_\theta $ over the teacher distribution $\pi_T $, under a uniform prior over the two models~\cite{gutmann2010noise}.
This also connects  to contrastive decoding~\cite{li2022contrastive}, which was applied at model {\textit{inference}} time  by amplifying the target model's log-probabilities relative to a weaker model.
In contrast, our formulation applies an analogous principle during \textit{training}, which uses the student-teacher ratio as a soft reweighting to prioritize learning in regions the student is likely to inhabit at inference time.

We demonstrate in Section~\draftblue{\ref{sec:experiments}} that this distribution correction improves distilled model performance beyond the instruction prompting baseline with a carefully designed reasoning template.
In contrast, several baseline distillation methods can underperform the base model that is prompted with the same reasoning instruction but without fine-tuning.


\vspace{-0.1in}
\section{Related Work}\vspace{-0.1in}
We study offline LLM reasoning distillation that tackles the teacher-student distribution drift on fixed offline training traces. This work resides at the intersection of knowledge distillation,  {post-training} optimization, and mathematical reasoning supervision.

\vspace{-0.1in}
\paragraph{Knowledge Distillation (KD).}
Knowledge Distillation was initially proposed for conventional AI models, in which a compact student model learns from a stronger teacher via softened predictive distributions or intermediate representations ~\cite{bucilua2006model,ba2014deep,hinton2015distilling,romero2015fitnets}.
This paradigm has been extended to LLMs, where intermediate rationales beyond final answers can be distilled from a strong teacher to a smaller student
~\cite{sanh2019distilbert,jiao2020tinybert,ho2023large,hsieh2023distilling}. Early reasoning-distillation studies such as \textit{Fine-tune-CoT}~\citep{ho2023large} and \textit{Distilling Step-by-Step}~\citep{hsieh2023distilling} established the value of teacher-generated rationales as supervision for smaller models. Subsequent work has studied how the quality, format, and composition of teacher traces affect distillation. Other studies have improved teacher-generated traces through rationale selection, filtering, structured explanations, and math-specific data curation~\cite{wang2023self,wang2023scott,mukherjee2023orca,luo2023wizardmath,yu2024metamath,toshniwal2024openmathinstruct,chen2025unveiling}.
These works motivated the use of high-quality offline teacher-generated reasoning traces. However, most methods still train the student model by imitating fixed teacher trajectories under
teacher-conditioned prefix distributions.
They therefore leave the training-inference distribution drift relatively underexplored: the student is optimized under teacher prefixes during training, but autoregresses on its own prefixes during inference~\citep{agarwal2024onpolicy,gu2024minillm,ko2024distillm,jung2025todi}.
%

\vspace{-0.05in}
\paragraph{ LLM Post-Training with On-policy vs. Offline Data.}

A closer line of work tackles this distribution shift more directly: Online RL and other self-improvement methods optimize on model-generated trajectories over fixed teacher traces. PPO-style RLHF methods~\citep{schulman2017proximal,ouyang2022training} and recent reasoning-oriented RL methods such as GRPO~\citep{shao2024deepseekmathpushinglimitsmathematical,guo2025deepseekr1} generate rollouts from the target model and update the model using outcome or process rewards. Related online or mixed-policy distillation methods, including \textit{GKD}~\citep{agarwal2024onpolicy} and \textit{CoTD-PO}~\citep{niu2025cotd}, also use student-conditioned or iteratively generated trajectories to better match inference-time behavior. These methods directly reduce the gap between training and inference distributions, but they require additional online sampling, substantially more training-time tokens, and often encounter low-quality student-generated traces early in training.
A complementary line of work focuses on redesigning the distillation directions. \textit{MiniLLM}~\citep{gu2024minillm},
\textit{DistiLLM}~\citep{ko2024distillm}, and \textit{ToDi}~\citep{jung2025todi}
modify autoregressive knowledge-distillation losses through reverse KL, skewed
KL, or token-wise divergence control, to better align training and generation
behavior.
Methods such as DistiLLM still require online student sampling to minimize the reverse KL divergence.
{Our method  resides in the purely offline setting}, but
focuses on the correction at a different level: {instead of changing the divergence measure,}
we adaptively reweight teacher-token supervision based on the
student--teacher distribution mismatch.
In this sense, our method changes how the offline teacher supervision is allocated, thus producing a distribution-corrected training signal while preserving the sample efficiency of fixed teacher traces. {Our formulation can be readily combined with SFT and other KL-based divergence measures.}
Compared with online methods such as GKD and CoTD-PO, our method remains fully offline without requiring student rollouts.
Prior offline distillation work, such as ToDi, has mainly been applied to the instruction-following domain; we studied \methodd's effect in mathematical reasoning distillation as an empirically more challenging domain (Sec~\ref{sec:exp-main}).

\vspace{-0.1in}
\paragraph{Math Reasoning Supervision and Evaluation.}
Mathematical reasoning also motivates fine-grained supervision: uniform imitation treats all teacher tokens equally, even though an early algebraic or logical error can propagate through an entire solution. This has motivated process-level supervision and verifier-style frameworks. Outcome-level verifiers and reward models judge complete solutions or final answers~\citep{cobbe2021training,shen-etal-2021-generate-rank}, while process-supervised methods provide finer-grained feedback on intermediate reasoning steps~\citep{uesato2022solvingmathwordproblems,lightman2023lets,wang2024mathshepherd}. Related correction-based frameworks further use teacher feedback to identify and revise erroneous reasoning traces~\citep{yang2025supercorrect}.

%
%
Our work is complementary to this literature: we design a chain-of-thought reasoning template to elicit strong reasoning traces from the teacher model without introducing a new verifier or search procedure.
We then move beyond uniform token imitation and use mismatch-aware reweighting to emphasize teacher supervision that better aligns with the student-conditioned inference distribution.
%


\providecolor{upgreen}{RGB}{46,125,50}
\providecolor{downred}{RGB}{183,28,28}
\providecommand{\gainup}[2]{#1{\scriptsize\,\textcolor{upgreen}{($\uparrow$#2)}}}
\providecommand{\gaindown}[2]{#1{\scriptsize\,\textcolor{gray}{($\downarrow$#2)}}}
\providecommand{\gainzero}[1]{#1{\scriptsize\,\textcolor{gray}{($\pm$0.00)}}}

\vspace{-0.1in}
\section{Experiments}\label{sec:experiments}\vspace{-0.1in}

We evaluate our method on mathematical reasoning benchmarks across diverse teacher-student configurations, varying model families, sizes, and tasks, with both in- and out-of-domain assessment and qualitative trace analysis.
We provide additional results on more experimental settings in the Appendix~\ref{app:audit-details}. Our main experiments were organized around four questions:

\textbf{Q1.} Does our offline distillation method improve the student model's reasoning compared to state-of-the-art offline distillation counterparts (Section~\ref{sec:exp-main})?

\textbf{Q2.} Is the distillation  benefit  consistent across different model families (Section~\ref{sec:exp-model-family})?

\textbf{Q3.} Does the method actually reduce teacher-student behavior divergence under student-generated prefixes during inference time (Section~\ref{sec:exp-prefix})?

\textbf{Q4.} Does the resulting student reasoning performance reflect higher quality in correctness, conciseness, and reliability (Section~\ref{sec:exp-quality})?

\textbf{Q5.} Why do we adopt the proposed transformations for stabilizing the distribution correction weights instead of alternative formulations (Section~\ref{sec:ab})?

\vspace{-0.1in}
\subsection{Experimental Setup: Baselines, Metrics, and Evaluation Protocols}
\label{sec:exp-setup}
\label{sec:eval_setup}
\vspace{-0.1in}
\paragraph{Models and Teacher Reasoning Instruction.}
We used Qwen2.5-7B-Instruct and Qwen2.5-14B-Instruct as teachers~\citep{qwen2024qwen25}, with Qwen2.5-3B-Instruct, Qwen2.5-1.5B-Instruct, and Llama-3.2-3B as
students.
Teacher traces were generated offline using a structured reasoning prompt (Appendix~\ref{app:prompt}) that elicits five stages: problem understanding, strategy selection, solution, verification, and final answer. We retained only traces where the teacher's final answer is correct for distillation training.
The same generation-and-filtering pipeline was used for both MATH and GSM8K training data.

\vspace{-0.1in}
\paragraph{Benchmarks.}
For in-domain evaluation, we distill on MATH~\citep{hendrycks2021math}
and GSM8K~\citep{cobbe2021training} training splits and evaluate on their
corresponding test sets, \eg using MATH500~\citep{lightman2023lets} for
MATH-trained checkpoints.
For out-of-domain (OOD) evaluation, models trained on MATH were further tested on OlympiadBench~\citep{he2024olympiadbench}\footnote{For OlympiadBench, we evaluated all performance on the non-combinatorics subset as all training corpora contain no combinatorics problems.},
Omni-Math~\citep{gao2024omnimath}, and AMC23, which do not appear in the training domain.
%


\vspace{-0.1in}
\paragraph{Baselines.}
We compared both fully offline distillation objectives and an online baseline.
\uline{SFT} uses the basic cross-entropy loss on teacher-generated reasoning
traces.
\uline{KL} uses KL divergence to measure teacher-student distribution
difference on the same offline traces, \uline{symmetric KL} combines forward KL and reverse KL divergence terms, following the broader line of LLM distillation work that studies forward/reverse KL combinations and bidirectional divergence objectives~\citep{jung2025todi,wu-etal-2025-rethinking}.
\uline{ToDi}~\citep{jung2025todi} is an offline distillation baseline.
\uline{GKD}~\citep{agarwal2024onpolicy} is an \textit{online} method reported for reference that relies on frequent sampling of student-generated traces for training.
For objectives where our correction is applicable, including SFT and KL, the \methodd-augmented variant applies the proposed token-level reweighting to the same fixed teacher traces. See Appendix \ref{app:kl-sigmoid-objective} for detailed formulation.
We also report the base model, the CoT-prompted student, and the CoT-prompted teacher as reference points for interpreting lower-bound, prompting-only, and teacher-level performance.
%

\vspace{-0.1in}
\paragraph{Evaluation metrics.}
\uline{\textit{Final-answer accuracy}} ($\uparrow$) is our primary metric computed using a mathematical-equivalence pipeline.
%
%
\uline{\textit{Distributional ExAccErr}} $\downarrow$: For divergence measurement, we evaluated the additional teacher-student
next-token distribution gap induced by conditioning on student-generated
prefixes rather than teacher prefixes. The initial definition was proposed in  MiniLLM~\citep{gu2024minillm}.
%
Appendix~\ref{app:exaccerr-definition} provides the detailed formulation.
\uline{\textit{Reasoning reliability:}} In Section \ref{sec:exp-quality}, we evaluated whether correct answers produced by our method are supported by stable reasoning quality using both automatic metrics and AI-as-Judge annotations.
%
See Appendix~\ref{app:audit-details} for more details.
%
%

\vspace{-0.1in}
\subsection{ Reasoning Accuracy Comparison} \label{sec:exp-main}
\label{sec:main_results}\vspace{-0.1in}

This subsection evaluates final-answer accuracy along two axes: (i) whether
\method improves fully offline teacher-trace distillation compared with prior art, and
(ii) how offline distillation methods compare  with  online training with student-rollouts such as GKD, and how they compare with tuning-free methods using a carefully designed reasoning instruction. The results lead to three main observations:

                     \begin{table*}[!tbp]
\vspace{-0.5em}
\centering
\scriptsize
\setlength{\tabcolsep}{3pt}
\renewcommand{\arraystretch}{0.90}\vspace{-0.1in}
\caption{\small{Distillation from Qwen2.5-7B-Instruct $\rightarrow$ Qwen2.5-3B-Instruct.  \method generally improves over offline baselines on both in-domain and out-of-domain tasks, while also outperforming the training-free instruction-prompting approach. "w/ CoT Instruct" denotes model performance instructed by our proposed reasoning template without training.
All entries are accuracy (\%).
Parenthesized metrics indicate performance changes introduced by \method relative to the corresponding base objective, where available.}}
\label{tab:qwen-7b-3b}
\begin{tabular*}{\textwidth}{@{\extracolsep{\fill}}lccccc@{}}
\toprule
\multicolumn{1}{c}{} & \multicolumn{2}{c}{\textbf{In-Domain Performance}} & \multicolumn{3}{c}{\textbf{Out of Domain Performance}} \\
\cmidrule(lr){2-3}\cmidrule(lr){4-6}
\textbf{Model / method} & \textbf{MATH500} & \textbf{GSM8K} & \textbf{OlympiadBench} & \textbf{Omni-Math} & \textbf{AMC23} \\
\midrule
\rowcolor{black!6}\multicolumn{6}{l}{\textit{Reference models}} \\
Qwen2.5-3B (base) & 42.60 & 55.19 & 25.19 & 17.50 & 32.50 \\
Qwen2.5-3B w/ CoT Instruct     & 60.20 & 81.12 & 29.81 & 28.50 & 35.00 \\
Qwen2.5-7B  & 55.80 & 69.07 & 34.04 & 27.00 & 35.00 \\
Qwen2.5-7B w/ CoT Instruct (teacher)   & 69.40 & 88.78 & 39.04 & 36.50 & 45.00 \\
\midrule
\cellcolor{black!6}GKD (online distillation)         & \underline{62.80} & 83.62 & \underline{33.00} & \underline{32.00} & 35.00 \\ \midrule
\rowcolor{black!6}\multicolumn{6}{l}{\textit{Offline distilled methods}} \\
SFT                  & 58.00 & 83.47 & 31.00 & 26.50 & 32.50 \\
SFT w/ \methodd          & \gainup{\textbf{62.60}}{4.60} & \gaindown{83.02}{0.45} & \gainup{\textbf{31.70}}{0.70} & \gainup{\textbf{29.50}}{3.00} & \gainup{\underline{\textbf{40.00}}}{7.50} \\
ToDi                 & 59.20 & 83.09 & 31.20 & 28.00 & 37.50 \\
KL                   & 59.00 & \underline{\textbf{83.93}} & 30.00 & 26.00 & 32.50 \\
Symmetric KL         & 59.20 & 83.55 & 28.65 & 28.50 & 37.50 \\
Symmetric KL w/ \method  & \gainup{61.80}{2.60} & \gainzero{83.55} & \gainup{30.38}{1.73} & \gainup{29.00}{0.50} & \gainzero{37.50} \\
\midrule
\bottomrule
\end{tabular*}
\vspace{-0.35em}
\vspace{-0.1in}
\end{table*}

First, {\method  yields the most consistent gains on MATH-style reasoning tasks}, where solutions typically require longer reasoning traces and prefix drift is more likely to affect generation.
As shown in Table \ref{tab:qwen-7b-3b}, on the in-domain MATH500 benchmark, \method improves the corresponding offline objectives in the main Qwen settings.
The gains are not limited to standard teacher-CoT SFT: our method  also improves the symmetric-KL objective in multiple settings, suggesting that the proposed correction can serve as a flexible reweighting mechanism for different offline distillation losses.

Comparisons with tuning-free prompting further highlight this effect.
In the 7B$\rightarrow$3B setting, SFT underperforms the instruction-prompted student on MATH500, while \method achieves performance beyond instruction prompting.
This indicates that direct behavior cloning on teacher traces may weaken the student's potential for reasoning elicitation, whereas our method allocates supervision more effectively to teacher tokens that remain compatible with the student's own generation behavior.

Results on GSM8K are less discriminative, likely because the task requires shorter reasoning traces and is less impacted by prefix distribution drift.
Strong baselines already plateau after training on traces generated by our reasoning instruction.
For example, in the 7B$\rightarrow$3B setting, SFT+\method is slightly below SFT on GSM8K, and symmetric KL+\method  is tied with symmetric KL, even though \method improves the corresponding MATH500 results and several OOD benchmarks.
We  consider GSM8K as a shorter-trace setting where the benefit of distribution correction is less pronounced.
This interpretation is consistent with our ablations in Section~\ref{sec:ab} and Appendix \ref{app:ablation}, where different transformations of the correction weight have more pronounced effects on longer MATH-style tasks, whereas GSM8K is less sensitive to a specific correction-weighting design.

Second, the improvements transfer to OOD competition-style benchmarks.
In the 7B$\rightarrow$3B setting, \method outperforms the corresponding offline objectives on several OOD evaluations, including OlympiadBench, Omni-Math, and AMC23.
These results suggest that distribution-corrected distillation improves generalization to related competition-style mathematical reasoning tasks.

Third, our method narrows the gap to online student-rollout distillation.
GKD remains stronger on some in-domain and OOD benchmarks, as expected, because it trains on student-conditioned online samples and directly observes the student's rollout distribution.
In contrast, \method  uses only fixed teacher traces, yet achieves stronger performance than offline baselines.
Concretely, \method{} generates no student-rollout tokens during training and uses only 3.12 GPU-hours, while GKD generates about 2.86M online student-rollout tokens and requires 49.56 GPU-hours, about 15.9$\times$ more compute.
This suggests that our distribution-corrected distillation can approach  the benefits of online training while preserving the sample efficiency of offline learning.

\begin{table*}[!tbp]\vspace{-0.1in}
\centering
\scriptsize
\setlength{\tabcolsep}{3pt}
\renewcommand{\arraystretch}{0.90}
\caption{\small{Accuracy (\%) of Student Models Distilled  from Qwen2.5-14B-Instruct. MATH500 and GSM8K reported performance of models distilled using MATH and GSM8K training data split, respectively.}}
\label{tab:qwen-14b-indomain}
\begin{tabular*}{\textwidth}{@{\extracolsep{\fill}}lcccc@{}}
\toprule
\multicolumn{1}{c}{} & \multicolumn{2}{c}{\textbf{14B $\rightarrow$ 3B}} & \multicolumn{2}{c}{\textbf{14B $\rightarrow$ 1.5B}} \\
\cmidrule(lr){2-3}\cmidrule(lr){4-5}
\textbf{Model / method} & \textbf{MATH500} & \textbf{GSM8K} & \textbf{MATH500} & \textbf{GSM8K} \\
\midrule
\rowcolor{black!6}\multicolumn{5}{l}{\textit{Reference models}} \\
Qwen2.5-3B (base)   & 42.60 & 55.19 & -- & -- \\
Qwen2.5-3B w/ CoT Instruct      & 60.20 & 81.12 & -- & -- \\
Qwen2.5-1.5B (base) & -- & --  & 19.80 & 21.76 \\
Qwen2.5-1.5B w/ CoT Instruct    & -- & -- & 43.60 & 50.34 \\
Qwen2.5-14B    & 56.40 & 71.19 & 56.40 & 71.19 \\
Qwen2.5-14B w/ CoT Instruct     & 71.20 & 90.07 & 71.20 & 90.07 \\
\midrule
\rowcolor{black!6}\multicolumn{5}{l}{\textit{Distilled student}} \\
SFT                             & 59.60 & 81.27 & 41.00 & 67.32 \\
SFT + \textbf{\methodd}          & \gainup{\underline{\textbf{61.80}}}{2.20} & \gainup{\underline{\textbf{82.79}}}{1.52} & \gainup{45.40}{4.40} & \gainup{69.75}{2.43} \\
ToDi                            & 60.40 & 82.34 & 45.40 & 69.29 \\
KL                              & 57.60 & 82.03 & 37.40 & \underline{\textbf{70.81}} \\
Symmetric KL                    & 61.40 & 81.80 & 44.20 & 69.75 \\
Symmetric KL + \textbf{\methodd}  & \gaindown{61.20}{0.20} & \gainup{82.49}{0.69} & \gainup{\underline{\textbf{50.20}}}{6.00} & \gainup{70.28}{0.53} \\
\bottomrule
\end{tabular*}
\vspace{-0.35em}
%
\vspace{-1.0em}
\end{table*}

\vspace{-0.1in}
\subsection{Robustness Across Model Distillation Settings} \label{sec:exp-model-family} \vspace{-0.1in}

\begin{wraptable}[15]{r}{0.5\textwidth}
\vspace{-0.1in}
\centering
\scriptsize
\setlength{\tabcolsep}{3pt}
\renewcommand{\arraystretch}{1}
\setlength{\abovecaptionskip}{1pt}
\setlength{\belowcaptionskip}{1pt}
\caption{\small{MATH500 accuracy for Qwen2.5 $\to$ Llama-3.2-3B distillation. \method consistently improves offline baselines across model families.}}%
\label{tab:cross_family}
\begin{tabular*}{\linewidth}{@{\extracolsep{\fill}}lcc@{}}

\toprule
\textbf{Method / Teacher Model} & \textbf{Qwen2.5-7B} & \textbf{Qwen2.5-14B} \\
\midrule
\rowcolor{black!6}\multicolumn{3}{@{}l}{\textit{Reference models}} \\
Llama direct & 22.60 & 22.60 \\
Llama CoT & 43.20 & 43.20 \\
Qwen 7B CoT & 69.40 & -- \\
Qwen 14B CoT & -- & 71.20 \\
\midrule
\rowcolor{black!6}\multicolumn{3}{@{}l}{\textit{Distilled student}} \\
SFT & 49.40 & 50.00 \\
SFT+\methodd & \gainup{\textbf{52.40}}{3.00} & \gainup{\textbf{52.00}}{2.00} \\
ToDi & 48.80 & 49.20 \\
KL & 50.80 & 50.20 \\
Sym. KL & 50.20 & 49.80 \\
Sym. KL+\methodd & \gainup{50.80}{0.60} & \gainup{51.00}{1.20} \\
\bottomrule
\end{tabular*}
\vspace{-0.6em}
\end{wraptable}
We further examined the robustness of different methods under varying teacher-student size configurations and model families.
Table~\ref{tab:qwen-14b-indomain} reports results using Qwen2.5-14B-Instruct as the teacher and Qwen2.5-3B-Instruct or Qwen2.5-1.5B-Instruct as the student, which are consistent with the main findings: our  distribution correction method generally improves offline distillation objectives across both student sizes, and  is most beneficial for longer mathematical reasoning traces, where teacher-student prefix mismatch has more room to affect generation.
For the 14B$\rightarrow$3B setting, SFT+\method improves over SFT on both MATH500 and GSM8K, with gains of 2.20 and 1.52 points, respectively.
For the smaller 14B$\rightarrow$1.5B setting, the gains are larger on MATH500.

To isolate the effect of the student model family, we also distilled a Llama-3.2-3B student from Qwen2.5 teachers using the same objectives as in the main experiments, and evaluated the resulting models on MATH500.
As shown in Table~\ref{tab:cross_family}, the cross-family results are consistent with the primary findings in Section~\ref{sec:main_results}: the \method-augmented objectives remain competitive with baselines that minimize teacher-student logit divergence and outperform the untuned base model with CoT instructions.
In both teacher settings, SFT+\method is the strongest distilled method, which improves over SFT by 3.0 and 2.0 points, respectively.
Consistent with Section~\ref{sec:main_results}, adding \method also improves the symmetric-KL objective in both settings.
 These results suggest that the proposed mismatch-aware reweighting remains effective across different distillation settings.

\vspace{-0.1in}
\subsection{Distribution Divergence Measurement}
\label{sec:exp-prefix}\vspace{-0.1in}

\begin{wrapfigure}[17]{r}{0.5\textwidth}
\vspace{-0.22in}
\centering
\includegraphics[width=1\linewidth]{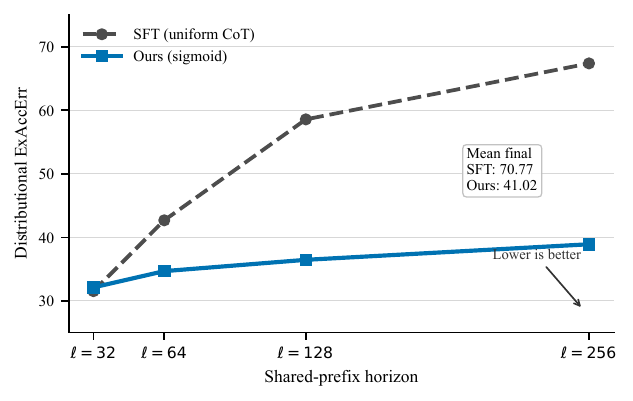}
\vspace{-0.3in}
\caption{\small{Comparing policy divergence  (ExAccErr):  A base model provides reasoning prefixes truncated at length $t \in \{32,64,128,256\}$.
Lower ExAccErr indicates less divergence in continuation behavior under student-generated prefixes.
\method remains more robust to prefix drift than  SFT as prefix length increases.
}}
\label{fig:prefix_diagnostics}
\end{wrapfigure}

To evaluate reasoning robustness under student-generated prefixes, we sampled prefixes from the  base student before training instead of from teacher traces.
Specifically, we randomly sampled 200 MATH test problems and let the base student model generate partial solutions.
From each partial solution, we truncated prefix horizons with $t \in \{32,64,128,256\}$ tokens.
Figure~\ref{fig:prefix_diagnostics} reports ExAccErr ($\downarrow$) for SFT and our method, where lower values indicate smaller divergence between the student model’s continuation answers and the expected answer distribution under the given prefixes. Specifically, each continuation answer is sampled from the trained student model conditioned on a provided prefix.

As the prefix horizon $t$ increases,  SFT exhibits a notably growing ExAccErr, which indicates that its behavior becomes increasingly unstable when conditioned on longer student-generated reasoning prefixes.
In contrast, \method keeps ExAccErr stable across prefix lengths.
This indicates that our proposed distribution correction improves robustness to prefix drift and reduces the amplification of errors introduced by student-generated intermediate reasoning.
Overall, this result shows that the benefit of \method is not limited to final-answer accuracy.
By reducing sensitivity to long student-generated prefixes, it better preserves reliable continuation behavior within support regions likely to be visited by student models.
\vspace{-0.12 in}
\subsection{Reasoning Quality Analysis}\label{sec:exp-quality}\vspace{-0.1 in}
In addition to the improved final-answer accuracy and reduced teacher-student distribution drift as shown above, we  examined whether performance gains of our method manifest in the generated reasoning traces themselves.
In particular, we investigated whether the correct answers are supported by more logically coherent and stable reasoning traces, and whether reducing the distribution drift in training alleviates typical generation failures such as repetition, post-answer continuation (\ie non-stop generation after providing a final answer), and inconsistent final
answers (\ie providing multiple distinct answers). For fair comparisons, we focused primarily on \textit{commonly-correct} examples where both SFT and \method produce the correct final answer.

\begin{wraptable}{r}{0.50\textwidth}
\vspace{-1.2 em}
\centering
\scriptsize
\setlength{\tabcolsep}{2pt}
\renewcommand{\arraystretch}{0.80}
\setlength{\abovecaptionskip}{0pt}
\setlength{\belowcaptionskip}{2pt}
\captionof{table}{Comparing reasoning-trace quality between SFT and \methodd. Lower values indicate more desirable traces.
}%
\label{tab:trace_metrics}
\begin{tabular*}{\linewidth}{@{\extracolsep{\fill}}p{0.24\linewidth}cccc@{}}
\toprule
& \multicolumn{2}{c}{\textbf{OlympiadBench}} 
& \multicolumn{2}{c}{\textbf{MATH}} \\
\cmidrule(lr){2-3}\cmidrule(lr){4-5}
\textbf{Metric} 
& \textbf{SFT} 
& \textbf{\method{}}
& \textbf{SFT} 
& \textbf{\method{}} \\
\midrule
Mean len. $\downarrow$    & 5328.1 & \textbf{1651.0} & 4383.0 & \textbf{1357.9} \\
Rep. 4-gram $\downarrow$  & 0.5603 & \textbf{0.3288} & 0.6514 & \textbf{0.2698} \\
Post-answer $\downarrow$  & 0.6000 & \textbf{0.1000} & 0.7900 & \textbf{0.1500} \\
Multi-answer $\downarrow$ & 0.5900 & \textbf{0.0000} & 0.7500 & \textbf{0.0000} \\
\bottomrule
\end{tabular*}
\vspace{-0.6em}
\end{wraptable}

\vspace{-0.1in}
\paragraph{Reasoning Trace Pattern Quality.}
To examine whether our method improves generation patterns, we compared \method with the SFT baseline using four automatic trace-quality metrics:
mean trace length in characters (\uline{Mean len}),
the fraction of repeated 4-grams in the generated trace (\uline{Rep. 4-gram}),
the rate of continuing generation after emitting a final answer (\uline{Post-answer}),
and the rate of producing multiple distinct final-answer statements within the same trace (\uline{Multi-answer}).
%
%
As shown in Table~\ref{tab:trace_metrics},
%
%
%
on MATH, our method decreases the mean trace length from 4383.0 to 1357.9 characters and the multiple-answer rate from 0.75 to 0, while also substantially lowering repetition and post-answer continuation rates.
A similar pattern also appears on OlympiadBench. Beyond improving final-answer accuracy, even within the \textit{commonly} correct subset, \method yields cleaner generation patterns by producing shorter traces with fewer repetitions, post-answer continuations, and multiple-answer failures.

\paragraph{Evaluation of Reasoning Trace Quality.}
\begin{wrapfigure}[15]{r}{0.32\textwidth}
  \vspace{-0.52in}
  \centering
  \includegraphics[width=\linewidth]{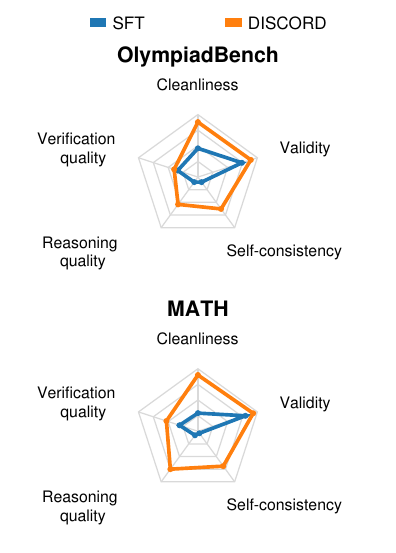}
  \vspace{-0.35in}
  \caption{\small{Comparing reasoning trace quality.}}
  \label{fig:table7-radar}
  \label{fig:trace_radar}
  \vspace{-0.4em}
\end{wrapfigure}
To complement the automatic metrics above, we further evaluated the generated traces using GPT-5.4 mini as a judge with a predefined rubric.
The rubric covers five dimensions: \uline{cleanliness}, \uline{validity} (\ie whether the reasoning is valid when the answer is correct), \uline{self-consistency}, \uline{reasoning quality}, and \uline{verification quality}. Appendix~\ref{app:trace-annotation-results} provides the definitions of each metric, the scoring protocol, and the full results.
As shown in Figure~\ref{fig:trace_radar}, these AI-as-judge annotation results are consistent with the evaluation using automatic metrics in Table \ref{tab:trace_metrics}.
%
Crucially, this suggests that the shorter responses produced by \method are not merely more concise
but also more logically coherent and internally consistent.
%
The detailed results further show that \method reduces the fraction
of traces with a correct final answer but flawed reasoning from 26\% to 11\% on OlympiadBench and from 20\% to 7\% on MATH.
In addition, \method receives higher
reasoning and verification quality scores.
These results together indicate a qualitative improvement in the reasoning traces underlying the final-answer accuracy gains.

\paragraph{Reasoning Trace Case Study.}
We further analyzed representative examples from one in-domain dataset, MATH, and one out-of-domain dataset, OlympiadBench. A concise case summary and representative trace analyses are provided in Appendix~\ref{app:reasoning-trace-case-study}.
The examples illustrate two qualitative patterns.
(1) First, in failure-recovery cases, where \method answers correctly while the SFT baseline fails (example A--D), our method more reliably preserves the underlying reasoning structure of the teacher solution, even when it does not exactly reproduce the teacher's surface-level formatting, such as the section headers: ``\texttt{PLAN}'', ``\texttt{VERIFICATION}'', etc.
Specifically, in the OlympiadBench divisor-counting problem, \method{} does not verbatim reproduce the teacher's verification section titled "\texttt{{VEriFICATION}}", but it keeps the teacher's core logic by using the correct factorization and checks the resulting answer consistently.
In contrast, the SFT baseline follows the expected template more closely, but its verification step uses an incorrect factorization and counting argument  that leads to the wrong answer.
(2) Second, in overlap-correct cases (example E--H), where both models reach the correct final answer, \method{} often produces a cleaner and more direct solution, with fewer unnecessary branches and more relevant checks. These observations are consistent with the aforementioned metrics: \method{} can improve reasoning trace reliability by preserving core reasoning content instead of  verbatim imitating the teacher's formatting.

Overall, our qualitative studies indicate that \method improves reasoning trace reliability beyond final-answer accuracy.
On commonly correct examples, our method yields  cleaner and more logically stable traces; in failure-recovery cases, it more often preserves the mathematical structure needed to reach the correct answer.
These findings imply that student-support-aware distribution correction strengthens offline reasoning distillation not by enforcing surface-level template imitation, but by reallocating teacher supervision toward reasoning behaviors that the student can more reliably carry through at inference time.
\subsection{Comparing Alternative Correction-Weight Designs  }\label{sec:ab} 
\begin{wraptable}[10]{r}{0.55\textwidth}
\vspace{-0.15in}
\centering
\scriptsize
\setlength{\tabcolsep}{2pt}
\renewcommand{\arraystretch}{0.88}
\setlength{\abovecaptionskip}{1pt}
\setlength{\belowcaptionskip}{1pt}
\caption{\small{Comparing the effects of varying correction-weight designs. Results are evaluated on MATH500  using  Qwen 7B $\to$ 3B distilled model. $\Delta_t=\log p_S(y_t\mid\cdot)-\log p_T(y_t\mid\cdot)$.}}
\label{tab:adv-transform}
\begin{tabular*}{\linewidth}{@{\extracolsep{\fill}}p{0.29\linewidth}p{0.49\linewidth}r@{}}
\toprule
\textbf{Variant} & \textbf{Weight / objective} & \textbf{Acc.} \\
\midrule
SFT & $w_t=1$ & 58.00 \\
ToDi & token-wise KL interpolation & 59.20 \\
\method, $\tau=1.0$ & $\sigma(\Delta_t/1.0)$ & \textbf{62.60} \\
\method, $\tau=2.0$ & $\sigma(\Delta_t/2.0)$ & 62.40 \\
\method, $\tau=4.0$ & $\sigma(\Delta_t/4.0)$ & 62.20 \\
Raw ratio & $\exp(\Delta_t)$ & 61.20 \\
ClipExp, $c=5.0$ & $\exp(\mathrm{clip}(\Delta_t,-5,5))$ & 62.00 \\
ReLU & $\max(\Delta_t,0)$ & 57.80 \\
\bottomrule
\end{tabular*}
\vspace{-0.35em}
\vspace{-1.0em}
\end{wraptable}
Table~\ref{tab:adv-transform} compares several correction-weight designs under the same prompt and evaluation setting.  The rows evaluate our proposed weighting from Eq.~(\ref{eq:kl-sigmoid-weight}) against clipped exponential weighting, the raw log-ratio, and a one-sided ReLU transform. Our formulation achieves the strongest result among the tested variants and inherits a theoretical interpretation grounded in noise-contrastive estimation.

\section{Conclusion and Future Work} \vspace{-0.1in}
We propose a simple yet effective offline distillation method for mitigating teacher–student distribution drift in tasks requiring long reasoning traces.
Our method keeps the training data fully offline but reweights teacher-token supervision based on a student–teacher support ratio, thereby prioritizing teacher behaviors that are more compatible with the student’s inference-time distribution.
Comprehensive evaluation on mathematical reasoning benchmarks shows that our method improves final-answer accuracy over strong offline baselines, narrows the gap to online student-rollout distillation, and yields more logical and higher-quality reasoning traces.
While our experiments focus on mathematical reasoning, the proposed method can be extended to other long-horizon generation settings, including code generation, scientific reasoning, and tool-use agents. Future work may include integrating our method with off-policy or online training to better balance data efficiency and distillation efficacy.
Another future direction is to decouple teacher trace generation from distribution correction scoring: when a trace-generating teacher is unavailable or unknown, a proxy reference model---potentially a weaker one---or an arbitrary policy may be used to approximate the support-ratio weights. This may preserve the core benefit of prioritizing student-reachable tokens while offering greater flexibility in the trade-off between teacher data quality and robustness to training-inference distribution shift.


\clearpage
\bibliographystyle{plainnat}
\bibliography{references}

\begin{thebibliography}{46}
\providecommand{\natexlab}[1]{#1}
\providecommand{\url}[1]{\texttt{#1}}
\expandafter\ifx\csname urlstyle\endcsname\relax
  \providecommand{\doi}[1]{doi: #1}\else
  \providecommand{\doi}{doi: \begingroup \urlstyle{rm}\Url}\fi

\bibitem[Agarwal et~al.(2024)Agarwal, Vieillard, Zhou, Stanczyk, Ramos~Garea,
  Geist, and Bachem]{agarwal2024onpolicy}
Rishabh Agarwal, Nino Vieillard, Yongchao Zhou, Piotr Stanczyk, Sabela
  Ramos~Garea, Matthieu Geist, and Olivier Bachem.
\newblock On-policy distillation of language models: Learning from
  self-generated mistakes.
\newblock In \emph{The Twelfth International Conference on Learning
  Representations}, 2024.
\newblock URL \url{https://openreview.net/forum?id=3zKtaqxLhW}.

\bibitem[Ba and Caruana(2014)]{ba2014deep}
Lei~Jimmy Ba and Rich Caruana.
\newblock Do deep nets really need to be deep?
\newblock In \emph{Advances in Neural Information Processing Systems},
  volume~27, pages 2654--2662, 2014.
\newblock URL
  \url{https://papers.nips.cc/paper/5484-do-deep-nets-really-need-to-be-deep}.

\bibitem[Bucilu{\u{a}} et~al.(2006)Bucilu{\u{a}}, Caruana, and
  Niculescu-Mizil]{bucilua2006model}
Cristian Bucilu{\u{a}}, Rich Caruana, and Alexandru Niculescu-Mizil.
\newblock Model compression.
\newblock In \emph{Proceedings of the 12th ACM SIGKDD International Conference
  on Knowledge Discovery and Data Mining}, pages 535--541. ACM, 2006.
\newblock \doi{10.1145/1150402.1150464}.

\bibitem[Cao and Kang(2025)]{cao2025llm}
Yihan Cao and Yanbin Kang.
\newblock On {LLM} knowledge distillation: A comparison between forward {KL}
  and reverse {KL}.
\newblock In \emph{The Fourth Blogpost Track at ICLR 2025}, 2025.
\newblock URL \url{https://openreview.net/forum?id=jGVCs8gomF}.

\bibitem[Chen et~al.(2025)Chen, Sun, Guo, Zhang, Chen, Sun, Su, Pan, Klakow,
  Li, and Shen]{chen2025unveiling}
Xinghao Chen, Zhijing Sun, Wenjin Guo, Miaoran Zhang, Yanjun Chen, Yirong Sun,
  Hui Su, Yijie Pan, Dietrich Klakow, Wenjie Li, and Xiaoyu Shen.
\newblock Unveiling the key factors for distilling chain-of-thought reasoning.
\newblock In \emph{Findings of the Association for Computational Linguistics:
  ACL 2025}, pages 15094--15119, 2025.
\newblock \doi{10.18653/v1/2025.findings-acl.782}.
\newblock URL \url{https://aclanthology.org/2025.findings-acl.782/}.

\bibitem[Cobbe et~al.(2021)Cobbe, Kosaraju, Bavarian, Chen, Jun, Kaiser,
  Plappert, Tworek, Hilton, Nakano, Hesse, and Schulman]{cobbe2021training}
Karl Cobbe, Vineet Kosaraju, Mohammad Bavarian, Mark Chen, Heewoo Jun, Lukasz
  Kaiser, Matthias Plappert, Jerry Tworek, Jacob Hilton, Reiichiro Nakano,
  Christopher Hesse, and John Schulman.
\newblock Training verifiers to solve math word problems.
\newblock \emph{arXiv preprint arXiv:2110.14168}, 2021.
\newblock URL \url{https://arxiv.org/abs/2110.14168}.

\bibitem[{DeepSeek-AI} et~al.(2025){DeepSeek-AI}, Guo, Yang, Zhang, Song,
  et~al.]{guo2025deepseekr1}
{DeepSeek-AI}, Daya Guo, Dejian Yang, Haowei Zhang, Junxiao Song, et~al.
\newblock {DeepSeek-R1}: Incentivizing reasoning capability in {LLM}s via
  reinforcement learning.
\newblock \emph{arXiv preprint arXiv:2501.12948}, 2025.
\newblock URL \url{https://arxiv.org/abs/2501.12948}.

\bibitem[Gao et~al.(2025)Gao, Song, Yang, Cai, Miao, Dong, Li, Ma, Chen, Xu,
  Tang, Wang, Zan, Quan, Zhang, Sha, Zhang, Ren, Liu, and
  Chang]{gao2024omnimath}
Bofei Gao, Feifan Song, Zhe Yang, Zefan Cai, Yibo Miao, Qingxiu Dong, Lei Li,
  Chenghao Ma, Liang Chen, Runxin Xu, Zhengyang Tang, Benyou Wang, Daoguang
  Zan, Shanghaoran Quan, Ge~Zhang, Lei Sha, Yichang Zhang, Xuancheng Ren,
  Tianyu Liu, and Baobao Chang.
\newblock {Omni-MATH}: A universal olympiad level mathematic benchmark for
  large language models.
\newblock In \emph{The Thirteenth International Conference on Learning
  Representations}, 2025.
\newblock URL
  \url{https://proceedings.iclr.cc/paper_files/paper/2025/hash/f9e1e8b56c7e363985ebeb0e9dd1a85c-Abstract-Conference.html}.

\bibitem[Gu et~al.(2024)Gu, Dong, Wei, and Huang]{gu2024minillm}
Yuxian Gu, Li~Dong, Furu Wei, and Minlie Huang.
\newblock {MiniLLM}: Knowledge distillation of large language models.
\newblock In \emph{International Conference on Learning Representations}, 2024.
\newblock URL \url{https://openreview.net/forum?id=5h0qf7IBZZ}.

\bibitem[Gutmann and Hyv{\"a}rinen(2010)]{gutmann2010noise}
Michael Gutmann and Aapo Hyv{\"a}rinen.
\newblock Noise-contrastive estimation: A new estimation principle for
  unnormalized statistical models.
\newblock In \emph{Proceedings of the Thirteenth International Conference on
  Artificial Intelligence and Statistics}, volume~9 of \emph{Proceedings of
  Machine Learning Research}, pages 297--304. PMLR, 2010.
\newblock URL \url{https://proceedings.mlr.press/v9/gutmann10a.html}.

\bibitem[He et~al.(2024)He, Luo, Bai, Hu, Thai, Shen, Hu, Han, Huang, Zhang,
  Liu, Qi, Liu, and Sun]{he2024olympiadbench}
Chaoqun He, Renjie Luo, Yuzhuo Bai, Shengding Hu, Zhen Thai, Junhao Shen, Jinyi
  Hu, Xu~Han, Yujie Huang, Yuxiang Zhang, Jie Liu, Lei Qi, Zhiyuan Liu, and
  Maosong Sun.
\newblock {OlympiadBench}: A challenging benchmark for promoting {AGI} with
  olympiad-level bilingual multimodal scientific problems.
\newblock In \emph{Proceedings of the 62nd Annual Meeting of the Association
  for Computational Linguistics (Volume 1: Long Papers)}, pages 3828--3850,
  2024.
\newblock \doi{10.18653/v1/2024.acl-long.211}.
\newblock URL \url{https://aclanthology.org/2024.acl-long.211/}.

\bibitem[Hendrycks et~al.(2021)Hendrycks, Burns, Kadavath, Arora, Basart, Tang,
  Song, and Steinhardt]{hendrycks2021math}
Dan Hendrycks, Collin Burns, Saurav Kadavath, Akul Arora, Steven Basart, Eric
  Tang, Dawn Song, and Jacob Steinhardt.
\newblock Measuring mathematical problem solving with the {MATH} dataset.
\newblock \emph{arXiv preprint arXiv:2103.03874}, 2021.
\newblock URL \url{https://arxiv.org/abs/2103.03874}.

\bibitem[Hinton et~al.(2015)Hinton, Vinyals, and Dean]{hinton2015distilling}
Geoffrey Hinton, Oriol Vinyals, and Jeff Dean.
\newblock Distilling the knowledge in a neural network.
\newblock \emph{arXiv preprint arXiv:1503.02531}, 2015.
\newblock \doi{10.48550/arXiv.1503.02531}.
\newblock URL \url{https://arxiv.org/abs/1503.02531}.

\bibitem[Ho et~al.(2023)Ho, Schmid, and Yun]{ho2023large}
Namgyu Ho, Laura Schmid, and Se-Young Yun.
\newblock Large language models are reasoning teachers.
\newblock In \emph{Proceedings of the 61st Annual Meeting of the Association
  for Computational Linguistics (Volume 1: Long Papers)}, pages 14852--14882,
  2023.
\newblock \doi{10.18653/v1/2023.acl-long.830}.
\newblock URL \url{https://aclanthology.org/2023.acl-long.830/}.

\bibitem[Hsieh et~al.(2023)Hsieh, Li, Yeh, Nakhost, Fujii, Ratner, Krishna,
  Lee, and Pfister]{hsieh2023distilling}
Cheng-Yu Hsieh, Chun-Liang Li, Chih-kuan Yeh, Hootan Nakhost, Yasuhisa Fujii,
  Alex Ratner, Ranjay Krishna, Chen-Yu Lee, and Tomas Pfister.
\newblock Distilling step-by-step! outperforming larger language models with
  less training data and smaller model sizes.
\newblock In \emph{Findings of the Association for Computational Linguistics:
  ACL 2023}, pages 8003--8017, 2023.
\newblock \doi{10.18653/v1/2023.findings-acl.507}.
\newblock URL \url{https://aclanthology.org/2023.findings-acl.507/}.

\bibitem[Jiao et~al.(2020)Jiao, Yin, Shang, Jiang, Chen, Li, Wang, and
  Liu]{jiao2020tinybert}
Xiaoqi Jiao, Yichun Yin, Lifeng Shang, Xin Jiang, Xiao Chen, Linlin Li, Fang
  Wang, and Qun Liu.
\newblock {TinyBERT}: Distilling {BERT} for natural language understanding.
\newblock In \emph{Findings of the Association for Computational Linguistics:
  EMNLP 2020}, pages 4163--4174. Association for Computational Linguistics,
  2020.
\newblock \doi{10.18653/v1/2020.findings-emnlp.372}.
\newblock URL \url{https://aclanthology.org/2020.findings-emnlp.372/}.

\bibitem[Jung et~al.(2025)Jung, Yoon, Kim, and Lee]{jung2025todi}
Seongryong Jung, Suwan Yoon, DongGeon Kim, and Hwanhee Lee.
\newblock Todi: Token-wise distillation via fine-grained divergence control.
\newblock In \emph{Proceedings of the 2025 Conference on Empirical Methods in
  Natural Language Processing}, pages 8078--8091, 2025.
\newblock \doi{10.18653/v1/2025.emnlp-main.409}.
\newblock URL \url{https://aclanthology.org/2025.emnlp-main.409/}.

\bibitem[Ko et~al.(2024)Ko, Kim, Chen, and Yun]{ko2024distillm}
Jongwoo Ko, Sungnyun Kim, Tianyi Chen, and Se-Young Yun.
\newblock {DistiLLM}: Towards streamlined distillation for large language
  models.
\newblock In \emph{Proceedings of the 41st International Conference on Machine
  Learning}, volume 235 of \emph{Proceedings of Machine Learning Research},
  pages 24872--24895, 2024.
\newblock URL \url{https://proceedings.mlr.press/v235/ko24c.html}.

\bibitem[Li et~al.(2023)Li, Holtzman, Fried, Liang, Eisner, Hashimoto,
  Zettlemoyer, and Lewis]{li2022contrastive}
Xiang~Lisa Li, Ari Holtzman, Daniel Fried, Percy Liang, Jason Eisner,
  Tatsunori~B Hashimoto, Luke Zettlemoyer, and Mike Lewis.
\newblock Contrastive decoding: Open-ended text generation as optimization.
\newblock In \emph{Proceedings of the 61st annual meeting of the association
  for computational linguistics (volume 1: Long papers)}, pages 12286--12312,
  2023.
\newblock \doi{10.18653/v1/2023.acl-long.687}.
\newblock URL \url{https://aclanthology.org/2023.acl-long.687/}.

\bibitem[Lightman et~al.(2024)Lightman, Kosaraju, Burda, Edwards, Baker, Lee,
  Leike, Schulman, Sutskever, and Cobbe]{lightman2023lets}
Hunter Lightman, Vineet Kosaraju, Yuri Burda, Harrison Edwards, Bowen Baker,
  Teddy Lee, Jan Leike, John Schulman, Ilya Sutskever, and Karl Cobbe.
\newblock Let's verify step by step.
\newblock In \emph{The Twelfth International Conference on Learning
  Representations}, 2024.
\newblock URL \url{https://openreview.net/forum?id=v8L0pN6EOi}.

\bibitem[Liu et~al.(2025)Liu, Bai, Lu, Sun, Kong, Wang, Shan, Jose, Liu, Wen,
  Yu, and Cao]{liu2025tisdpo}
Aiwei Liu, Haoping Bai, Zhiyun Lu, Yanchao Sun, Xiang Kong, Xiaoming~Simon
  Wang, Jiulong Shan, Albin~Madappally Jose, Xiaojiang Liu, Lijie Wen,
  Philip~S. Yu, and Meng Cao.
\newblock {TIS}-{DPO}: Token-level importance sampling for direct preference
  optimization with estimated weights.
\newblock In \emph{The Thirteenth International Conference on Learning
  Representations}, 2025.
\newblock URL \url{https://openreview.net/forum?id=oF6e2WwxX0}.

\bibitem[Luo et~al.(2023)Luo, Sun, Xu, Zhao, Lou, Tao, Geng, Lin, Chen, and
  Zhang]{luo2023wizardmath}
Haipeng Luo, Qingfeng Sun, Can Xu, Pu~Zhao, Jianguang Lou, Chongyang Tao, Xiubo
  Geng, Qingwei Lin, Shifeng Chen, and Dongmei Zhang.
\newblock {WizardMath}: Empowering mathematical reasoning for large language
  models via reinforced evol-instruct.
\newblock \emph{arXiv preprint arXiv:2308.09583}, 2023.
\newblock URL \url{https://arxiv.org/abs/2308.09583}.

\bibitem[Mukherjee et~al.(2023)Mukherjee, Mitra, Jawahar, Agarwal, Palangi, and
  Awadallah]{mukherjee2023orca}
Subhabrata Mukherjee, Arindam Mitra, Ganesh Jawahar, Sahaj Agarwal, Hamid
  Palangi, and Ahmed Awadallah.
\newblock {Orca}: Progressive learning from complex explanation traces of
  {GPT-4}.
\newblock \emph{arXiv preprint arXiv:2306.02707}, 2023.
\newblock URL \url{https://arxiv.org/abs/2306.02707}.

\bibitem[Niu et~al.(2025)Niu, Sun, Zhao, Chen, Bai, Zhang, Yuan, and
  Wang]{niu2025cotd}
Lujie Niu, Haochen Sun, Fangkun Zhao, Sheng Chen, Zimeng Bai, Jiawei Zhang,
  Caixia Yuan, and Xiaojie Wang.
\newblock {CoTD-PO}: Chain-of-thought distillation with preference
  optimization.
\newblock In \emph{Findings of the Association for Computational Linguistics:
  EMNLP 2025}, pages 19975--19986, 2025.
\newblock \doi{10.18653/v1/2025.findings-emnlp.1087}.
\newblock URL \url{https://aclanthology.org/2025.findings-emnlp.1087/}.

\bibitem[Ouyang et~al.(2022)Ouyang, Wu, Jiang, Almeida, Wainwright, Mishkin,
  Zhang, Agarwal, Slama, Ray, Schulman, Hilton, Kelton, Miller, Simens, Askell,
  Welinder, Christiano, Leike, and Lowe]{ouyang2022training}
Long Ouyang, Jeffrey Wu, Xu~Jiang, Diogo Almeida, Carroll~L. Wainwright, Pamela
  Mishkin, Chong Zhang, Sandhini Agarwal, Katarina Slama, Alex Ray, John
  Schulman, Jacob Hilton, Fraser Kelton, Luke Miller, Maddie Simens, Amanda
  Askell, Peter Welinder, Paul Christiano, Jan Leike, and Ryan Lowe.
\newblock Training language models to follow instructions with human feedback.
\newblock In \emph{Advances in Neural Information Processing Systems},
  volume~35, pages 27730--27744, 2022.
\newblock URL
  \url{https://proceedings.neurips.cc/paper_files/paper/2022/hash/b1efde53be364a73914f58805a001731-Abstract-Conference.html}.

\bibitem[{Qwen Team}(2024)]{qwen2024qwen25}
{Qwen Team}.
\newblock {Qwen2.5} technical report.
\newblock \emph{arXiv preprint arXiv:2412.15115}, 2024.
\newblock URL \url{https://arxiv.org/abs/2412.15115}.

\bibitem[Romero et~al.(2015)Romero, Ballas, Ebrahimi~Kahou, Chassang, Gatta,
  and Bengio]{romero2015fitnets}
Adriana Romero, Nicolas Ballas, Samira Ebrahimi~Kahou, Antoine Chassang, Carlo
  Gatta, and Yoshua Bengio.
\newblock {FitNets}: Hints for thin deep nets.
\newblock In \emph{International Conference on Learning Representations}, 2015.
\newblock URL \url{https://arxiv.org/abs/1412.6550}.

\bibitem[Sanh et~al.(2019)Sanh, Debut, Chaumond, and Wolf]{sanh2019distilbert}
Victor Sanh, Lysandre Debut, Julien Chaumond, and Thomas Wolf.
\newblock {DistilBERT}, a distilled version of {BERT}: Smaller, faster, cheaper
  and lighter.
\newblock \emph{arXiv preprint arXiv:1910.01108}, 2019.
\newblock URL \url{https://arxiv.org/abs/1910.01108}.

\bibitem[Schulman et~al.(2017)Schulman, Wolski, Dhariwal, Radford, and
  Klimov]{schulman2017proximal}
John Schulman, Filip Wolski, Prafulla Dhariwal, Alec Radford, and Oleg Klimov.
\newblock Proximal policy optimization algorithms.
\newblock \emph{arXiv preprint arXiv:1707.06347}, 2017.
\newblock URL \url{https://arxiv.org/abs/1707.06347}.

\bibitem[Shao et~al.(2024)Shao, Wang, Zhu, Xu, Song, Bi, Zhang, Zhang, Li, Wu,
  and Guo]{shao2024deepseekmathpushinglimitsmathematical}
Zhihong Shao, Peiyi Wang, Qihao Zhu, Runxin Xu, Junxiao Song, Xiao Bi, Haowei
  Zhang, Mingchuan Zhang, Y.~K. Li, Y.~Wu, and Daya Guo.
\newblock Deepseekmath: Pushing the limits of mathematical reasoning in open
  language models, 2024.
\newblock URL \url{https://arxiv.org/abs/2402.03300}.

\bibitem[Shen et~al.(2021)Shen, Yin, Li, Shang, Jiang, Zhang, and
  Liu]{shen-etal-2021-generate-rank}
Jianhao Shen, Yichun Yin, Lin Li, Lifeng Shang, Xin Jiang, Ming Zhang, and Qun
  Liu.
\newblock Generate {\&} rank: A multi-task framework for math word problems.
\newblock In Marie-Francine Moens, Xuanjing Huang, Lucia Specia, and Scott
  Wen-tau Yih, editors, \emph{Findings of the Association for Computational
  Linguistics: EMNLP 2021}, pages 2269--2279, Punta Cana, Dominican Republic,
  November 2021. Association for Computational Linguistics.
\newblock \doi{10.18653/v1/2021.findings-emnlp.195}.
\newblock URL \url{https://aclanthology.org/2021.findings-emnlp.195/}.

\bibitem[Sutton et~al.(1999)Sutton, McAllester, Singh, and
  Mansour]{sutton1999policy}
Richard~S. Sutton, David McAllester, Satinder Singh, and Yishay Mansour.
\newblock Policy gradient methods for reinforcement learning with function
  approximation.
\newblock \emph{Advances in Neural Information Processing Systems}, 12, 1999.
\newblock URL
  \url{https://papers.nips.cc/paper/1713-policy-gradient-methods-for-reinforcement-learning-with-function-approximation}.

\bibitem[Toshniwal et~al.(2024)Toshniwal, Moshkov, Narenthiran, Gitman, Jia,
  and Gitman]{toshniwal2024openmathinstruct}
Shubham Toshniwal, Ivan Moshkov, Sean Narenthiran, Daria Gitman, Fei Jia, and
  Igor Gitman.
\newblock {OpenMathInstruct-1}: A 1.8 million math instruction tuning dataset.
\newblock In \emph{Advances in Neural Information Processing Systems},
  volume~37, 2024.
\newblock \doi{10.52202/079017-1096}.
\newblock URL
  \url{https://papers.nips.cc/paper_files/paper/2024/hash/3d5aa9a7ce28cdc710fbd044fd3610f3-Abstract-Datasets_and_Benchmarks_Track.html}.

\bibitem[Uesato et~al.(2022)Uesato, Kushman, Kumar, Song, Siegel, Wang,
  Creswell, Irving, and Higgins]{uesato2022solvingmathwordproblems}
Jonathan Uesato, Nate Kushman, Ramana Kumar, Francis Song, Noah Siegel, Lisa
  Wang, Antonia Creswell, Geoffrey Irving, and Irina Higgins.
\newblock Solving math word problems with process- and outcome-based feedback,
  2022.
\newblock URL \url{https://arxiv.org/abs/2211.14275}.

\bibitem[Wang et~al.(2023{\natexlab{a}})Wang, Wang, Li, Gao, Yin, and
  Ren]{wang2023scott}
Peifeng Wang, Zhengyang Wang, Zheng Li, Yifan Gao, Bing Yin, and Xiang Ren.
\newblock {SCOTT}: Self-consistent chain-of-thought distillation.
\newblock In \emph{Proceedings of the 61st Annual Meeting of the Association
  for Computational Linguistics (Volume 1: Long Papers)}, pages 5546--5558.
  Association for Computational Linguistics, 2023{\natexlab{a}}.
\newblock \doi{10.18653/v1/2023.acl-long.304}.
\newblock URL \url{https://aclanthology.org/2023.acl-long.304/}.

\bibitem[Wang et~al.(2024)Wang, Li, Shao, Xu, Dai, Li, Chen, Wu, and
  Sui]{wang2024mathshepherd}
Peiyi Wang, Lei Li, Zhihong Shao, Runxin Xu, Damai Dai, Yifei Li, Deli Chen,
  Yu~Wu, and Zhifang Sui.
\newblock {Math-Shepherd}: Verify and reinforce {LLM}s step-by-step without
  human annotations.
\newblock In \emph{Proceedings of the 62nd Annual Meeting of the Association
  for Computational Linguistics (Volume 1: Long Papers)}, pages 9426--9439,
  2024.
\newblock \doi{10.18653/v1/2024.acl-long.510}.
\newblock URL \url{https://aclanthology.org/2024.acl-long.510/}.

\bibitem[Wang et~al.(2023{\natexlab{b}})Wang, Wei, Schuurmans, Le, Chi, Narang,
  Chowdhery, and Zhou]{wang2023selfconsistency}
Xuezhi Wang, Jason Wei, Dale Schuurmans, Quoc~V. Le, Ed~H. Chi, Sharan Narang,
  Aakanksha Chowdhery, and Denny Zhou.
\newblock Self-consistency improves chain of thought reasoning in language
  models.
\newblock In \emph{International Conference on Learning Representations},
  2023{\natexlab{b}}.
\newblock URL \url{https://openreview.net/forum?id=1PL1NIMMrw}.

\bibitem[Wang et~al.(2023{\natexlab{c}})Wang, Kordi, Mishra, Liu, Smith,
  Khashabi, and Hajishirzi]{wang2023self}
Yizhong Wang, Yeganeh Kordi, Swaroop Mishra, Alisa Liu, Noah~A. Smith, Daniel
  Khashabi, and Hannaneh Hajishirzi.
\newblock Self-instruct: Aligning language models with self-generated
  instructions.
\newblock In \emph{Proceedings of the 61st Annual Meeting of the Association
  for Computational Linguistics (Volume 1: Long Papers)}, pages 13484--13508.
  Association for Computational Linguistics, 2023{\natexlab{c}}.
\newblock \doi{10.18653/v1/2023.acl-long.754}.
\newblock URL \url{https://aclanthology.org/2023.acl-long.754/}.

\bibitem[Wei et~al.(2022)Wei, Wang, Schuurmans, Bosma, Chi, Le, and
  Zhou]{wei2022chain}
Jason Wei, Xuezhi Wang, Dale Schuurmans, Maarten Bosma, Ed~H. Chi, Quoc~V. Le,
  and Denny Zhou.
\newblock Chain-of-thought prompting elicits reasoning in large language
  models.
\newblock In \emph{Advances in Neural Information Processing Systems},
  volume~35, pages 24824--24837, 2022.
\newblock URL
  \url{https://proceedings.neurips.cc/paper_files/paper/2022/hash/9d5609613524ecf4f15af0f7b31abca4-Abstract-Conference.html}.

\bibitem[Wu et~al.(2025)Wu, Tao, Wang, Yang, Zhao, and
  Wong]{wu-etal-2025-rethinking}
Taiqiang Wu, Chaofan Tao, Jiahao Wang, Runming Yang, Zhe Zhao, and Ngai Wong.
\newblock Rethinking {K}ullback-{L}eibler divergence in knowledge distillation
  for large language models.
\newblock In \emph{Proceedings of the 31st International Conference on
  Computational Linguistics}, pages 5737--5755, Abu Dhabi, UAE, January 2025.
  Association for Computational Linguistics.
\newblock URL \url{https://aclanthology.org/2025.coling-main.383/}.

\bibitem[Yang et~al.(2025)Yang, Yu, Zhang, Xu, Gonzalez, CUI, and
  YAN]{yang2025supercorrect}
Ling Yang, Zhaochen Yu, Tianjun Zhang, Minkai Xu, Joseph~E. Gonzalez, Bin CUI,
  and Shuicheng YAN.
\newblock Supercorrect: Advancing small {LLM} reasoning with thought template
  distillation and self-correction.
\newblock In \emph{The Thirteenth International Conference on Learning
  Representations}, 2025.
\newblock URL \url{https://openreview.net/forum?id=PyjZO7oSw2}.

\bibitem[Yang et~al.(2026)Yang, Lin, Liu, Tian, Liu, and
  Zhang]{yang2026tokenimportance}
Ning Yang, Hai Lin, Yibo Liu, Baoliang Tian, Guoqing Liu, and Haijun Zhang.
\newblock Token-importance guided direct preference optimization.
\newblock In \emph{The Fourteenth International Conference on Learning
  Representations}, 2026.
\newblock URL \url{https://openreview.net/forum?id=cMEnMVvMw9}.

\bibitem[Yoon et~al.(2024)Yoon, Yoon, Eom, Han, Nam, Jo, On, Hasegawa-Johnson,
  Kim, and Yoo]{yoon-etal-2024-tlcr}
Eunseop Yoon, Hee~Suk Yoon, SooHwan Eom, Gunsoo Han, Daniel Nam, Daejin Jo,
  Kyoung-Woon On, Mark Hasegawa-Johnson, Sungwoong Kim, and Chang Yoo.
\newblock {TLCR}: Token-level continuous reward for fine-grained reinforcement
  learning from human feedback.
\newblock In Lun-Wei Ku, Andre Martins, and Vivek Srikumar, editors,
  \emph{Findings of the Association for Computational Linguistics: ACL 2024},
  pages 14969--14981, Bangkok, Thailand, August 2024. Association for
  Computational Linguistics.
\newblock \doi{10.18653/v1/2024.findings-acl.889}.
\newblock URL \url{https://aclanthology.org/2024.findings-acl.889/}.

\bibitem[Yu et~al.(2024)Yu, Jiang, Shi, Yu, Liu, Zhang, Kwok, Li, Weller, and
  Liu]{yu2024metamath}
Longhui Yu, Weisen Jiang, Han Shi, Jincheng Yu, Zhengying Liu, Yu~Zhang,
  James~T. Kwok, Zhenguo Li, Adrian Weller, and Weiyang Liu.
\newblock {MetaMath}: Bootstrap your own mathematical questions for large
  language models.
\newblock In \emph{International Conference on Learning Representations}, 2024.
\newblock URL \url{https://openreview.net/forum?id=N8N0hgNDRt}.

\bibitem[Yu et~al.(2025)Yu, Zhang, Zhu, Yuan, Zuo, Yue, Dai, Fan, Liu, Liu,
  et~al.]{yu2025dapo}
Qiying Yu, Zheng Zhang, Ruofei Zhu, Yufeng Yuan, Xiaochen Zuo, Yu~Yue, Weinan
  Dai, Tiantian Fan, Gaohong Liu, Lingjun Liu, et~al.
\newblock Dapo: An open-source llm reinforcement learning system at scale.
\newblock \emph{arXiv preprint arXiv:2503.14476}, 2025.
\newblock \doi{10.48550/arXiv.2503.14476}.
\newblock URL \url{https://arxiv.org/abs/2503.14476}.

\bibitem[Zeng et~al.(2024)Zeng, Liu, Ma, Yang, Zhang, and Wang]{zeng2024tdpo}
Yongcheng Zeng, Guoqing Liu, Weiyu Ma, Ning Yang, Haifeng Zhang, and Jun Wang.
\newblock Token-level direct preference optimization.
\newblock In Ruslan Salakhutdinov, Zico Kolter, Katherine Heller, Adrian
  Weller, Nuria Oliver, Jonathan Scarlett, and Felix Berkenkamp, editors,
  \emph{Proceedings of the 41st International Conference on Machine Learning},
  volume 235 of \emph{Proceedings of Machine Learning Research}, pages
  58348--58365. PMLR, 21--27 Jul 2024.
\newblock URL \url{https://proceedings.mlr.press/v235/zeng24c.html}.

\end{thebibliography}
\clearpage
\appendix
\raggedbottom
\section*{Appendix}
\section{\method{} Temperature Tuning}\label{app:ablation}
Table~\ref{tab:appendix-discord-tau} reports the sigmoid temperatures used for
the DisCORD-augmented rows in the main result tables. We tune this temperature
within each teacher--student setting and base objective, rather than using a
single global value across all experiments. This is because the scale of the
student--teacher log-probability gap can change with teacher size, student
capacity, dataset, and whether DisCORD is applied on top of SFT or symmetric
KL. For MATH-trained checkpoints, the same temperature is used for the MATH500
in-domain evaluation and for the OOD evaluations, since the OOD benchmarks reuse
the MATH-trained models. GSM8K results are listed separately because they come
from independently trained GSM8K checkpoints.
\begin{center}
\footnotesize
\setlength{\tabcolsep}{5pt}
\renewcommand{\arraystretch}{1.08}
\captionof{table}{
Sigmoid temperatures used by the DisCORD rows in the main-result tables.
MATH-trained checkpoints are evaluated on MATH500 and also reused for the OOD
benchmarks; GSM8K numbers come from separately trained GSM8K checkpoints.
}
\label{tab:appendix-discord-tau}
\begin{tabular*}{\textwidth}{@{\extracolsep{\fill}}llcccc@{}}
\toprule
\textbf{Teacher $\rightarrow$ student}
& \textbf{Base objective}
& \textbf{MATH $\tau$}
& \textbf{MATH500}
& \textbf{GSM8K $\tau$}
& \textbf{GSM8K} \\
\midrule
Qwen2.5-7B $\rightarrow$ Qwen2.5-3B
& SFT
& 1.0 & 62.60
& 8.0 & 83.02 \\

Qwen2.5-7B $\rightarrow$ Qwen2.5-3B
& symmetric KL
& 2.0 & 61.80
& 8.0 & 83.55 \\

Qwen2.5-14B $\rightarrow$ Qwen2.5-3B
& SFT
& 4.0 & 61.80
& 4.0 & 82.79 \\

Qwen2.5-14B $\rightarrow$ Qwen2.5-3B
& symmetric KL
& 8.0 & 61.20
& 8.0 & 82.49 \\

Qwen2.5-14B $\rightarrow$ Qwen2.5-1.5B
& SFT
& 4.0 & 45.40
& 4.0 & 69.75 \\

Qwen2.5-14B $\rightarrow$ Qwen2.5-1.5B
& symmetric KL
& 1.0 & 50.20
& 1.0 & 70.28 \\
\bottomrule
\end{tabular*}
\end{center}

\section{Teacher Chain-of-Thought Prompt}
\label{app:teacher-prompt}
\label{app:prompt}

We designed this teacher CoT prompt ourselves to elicit structured, self-verifying
reasoning with explicit understanding, strategy, solution, and verification stages.
Following established prompt design principles such as role specification,
explicit output structure, and self-consistency checks \cite{wei2022chain,wang2023selfconsistency}, we refined the template on a held-out sample of training questions and did not tune the prompt
against the evaluation benchmarks.


\begin{tcolorbox}[colback=white,colframe=blue!75!black]
\begingroup\scriptsize
\begin{verbatim}
<role>
You are a distinguished mathematics researcher and olympiad problem solver
with expertise across all mathematical domains, from elementary algebra to
advanced proof-based mathematics. You approach each problem with the rigor of
peer-reviewed research and the creativity of competitive mathematics.
</role>

<instructions>
You will be presented with mathematical problems of varying difficulty. Your
goal is to provide solutions with complete transparency of reasoning, catching
and correcting any errors through systematic verification. If your initial
approach does not work, try alternative methods.

For every problem, respond using EXACTLY the following sections and order:
1. <understanding>...</understanding>
2. <strategy>...</strategy>
3. <solution>...</solution>
4. <verification>...</verification>
5. Answer: \boxed{FINAL_ANSWER}

## 1. UNDERSTAND
<understanding>
- Identify what is given and what needs to be found
- Note any constraints or special conditions
</understanding>

## 2. PLAN
<strategy>
- Outline your solution approach step-by-step
- Identify which mathematical techniques/theorems you will use
- Consider if there are multiple viable approaches
- Explain why you chose this particular strategy
</strategy>
\end{verbatim}
\endgroup
\end{tcolorbox}

\begin{tcolorbox}[colback=white,colframe=blue!75!black]
\begingroup\scriptsize
\begin{verbatim}
## 3. SOLVE
<solution>
Execute your plan with complete transparency:
- Show every calculation and transformation
- Justify each step with mathematical reasoning
- State which theorems/properties you are applying
- If you get stuck, explicitly note it and try a different approach
</solution>

## 4. VERIFY
<verification>
- Check your solution rigorously
- Substitute your answer back into the original problem
- Verify it satisfies all given conditions
- Review each step for logical errors or computational mistakes

If you find ANY error:
- Clearly state what the error was
- Explain why it occurred
- Show the corrected step
- Re-verify the corrected solution
</verification>

## 5. FINAL ANSWER
Present your final answer in the following format:
Answer: \boxed{FINAL_ANSWER}
</instructions>
\end{verbatim}
\endgroup
\end{tcolorbox}

\section{Training and Evaluation Audit Details}
\label{app:audit-details}

This section collects the implementation details used to make the experimental
comparisons auditable. The main comparison is objective-level: within each
teacher--student--dataset setting, all offline objectives are trained on the
same retained teacher-generated trace corpus and differ only in the loss or
token weighting applied to that corpus.

\paragraph{Training corpus construction.}
For MATH-family experiments, teacher traces are generated from the MATH training
split ($7{,}500$ source problems). For GSM8K experiments, teacher traces are
generated from the GSM8K training split ($7{,}473$ source problems). Each source
problem is prompted once with the structured teacher prompt in
Appendix~\ref{app:prompt}; the generated trace is then filtered by the same
final-answer equivalence judge used in evaluation. We retain only
teacher-correct traces for distillation. The retained subset, retained count,
and average teacher-trace length are fixed before student training and are
shared by all offline objectives within the corresponding teacher--student
setting. Cross-family Llama experiments reuse the same Qwen2.5 teacher-trace
construction pipeline as the Qwen-student experiments.

\paragraph{Offline Training Protocol and hyperparameters.}Within a teacher--student setting and data split, the fully offline methods
(SFT, KL, symmetric KL, ToDi, and \method{}-augmented variants) are trained on
the same retained teacher-correct examples and use the same offline training
recipe: AdamW with learning rate $3\times 10^{-6}$, weight decay $0.01$,
cosine decay with $5\%$ warmup, gradient clipping at norm $1.0$, bfloat16
mixed precision, maximum sequence length 2048, random seed 42, and effective
batch size 16. For MATH-family runs, this corresponds to micro-batch size 1
with gradient accumulation 16; for GSM8K, micro-batch size 2 with gradient
accumulation 8. The number of epochs is fixed within each matched setting and
base objective, so changes in Table~\ref{tab:qwen-7b-3b},
Table~\ref{tab:qwen-14b-indomain}, and Table~\ref{tab:cross_family} are not
attributable to different amounts of offline teacher data or different offline
training-token budgets. Hyperparameters that define an objective itself, such
as the KL/ToDi temperature or the sigmoid temperature for \method{}, are
selected per setting and base objective and are reported in the corresponding
table captions or rows when they differ.

\paragraph{Online Protocol and hyperparameters.}GKD is reported separately because it is not a compute-matched offline
baseline. It uses student rollouts during training and follows its own
on-policy recipe: AdamW with learning rate $3\times 10^{-6}$, weight decay
$0.01$, cosine decay with $5\%$ warmup, gradient clipping at norm $1.0$,
maximum sequence length 2048, effective batch size 16, $\beta=0.5$ following the original paper setting, and one
training epoch in the reported runs, with the on-policy mixing parameter
$\lambda$ specified by the corresponding run. We therefore treat GKD as an
online reference rather than as evidence about the matched offline
training-token budget.

\paragraph{Evaluation protocol.}
All methods in a setting use the same evaluation prompt, greedy decoding
protocol, and maximum generation length. The generation caps are 2048 new tokens
for GSM8K, MATH500, and 4096 new tokens for AMC23, Omni-Math and
OlympiadBench. Final-answer correctness is judged by GPT-5.4-mini using the
same mathematical-equivalence prompt for every method. The judge is not given
the method identity and is instructed to ignore reasoning quality and compare
only the final answer. If a solution contains multiple candidate answers, the
judge follows the explicitly marked final answer, such as one introduced by
``final'', ``answer'', ``therefore'', or \verb|\boxed{}|. The full judge prompt
appears in Appendix~\ref{app:evaluation-details}.

\paragraph{Compute resources.}
Experiments were run on SLURM-managed GPU clusters. Unless otherwise noted,
training and evaluation jobs used NVIDIA A100 40GB GPUs; GPU-hours are reported
as the number of allocated GPUs multiplied by wall-clock runtime. In the main
7B$\rightarrow$3B comparison, the \method{} run uses no online student-rollout
tokens and requires 3.12 GPU-hours, whereas the GKD online baseline generates
about 2.86M student-rollout tokens and requires 49.56 GPU-hours. The reported
GPU-hour comparison covers training compute for the matched distillation
setting; teacher-trace generation, benchmark evaluation, and GPT-5.4-mini judge
calls are one-time preprocessing or evaluation costs shared across methods
within each setting.

\section{Evaluation Details and Correctness Judge}
\label{app:evaluation-details}

Main in-domain evaluation uses MATH500~\citep{lightman2023lets} for
MATH-trained checkpoints ($n=500$) and GSM8K~\citep{cobbe2021training} for
GSM8K-trained checkpoints ($n=1{,}319$). OOD evaluation uses the
non-combinatorics portion of OlympiadBench \texttt{OE\_TO\_maths\_en\_COMP}
($n=520$ after removing combinatorics problems), the fixed Omni-Math D3/D4
paper subset~\citep{gao2024omnimath} ($n=200$), and AMC23 ($n=40$). We exclude
OlympiadBench combinatorics problems because the MATH teacher-trace training
corpus used here contains no combinatorics examples.

All distilled checkpoints are evaluated with the same finetuning-style prompt
template and greedy decoding. We use generation caps of 2048 new tokens for
GSM8K, MATH500, and AMC23, and 4096 new tokens for Omni-Math and
OlympiadBench.

We judge final-answer correctness with GPT-5.4-mini. The judge is instructed
to ignore reasoning quality and compare only the final answer by mathematical
equivalence. The prompt template is as below:

\begin{tcolorbox}[colback=white,colframe=blue!75!black]
\begingroup\scriptsize
\begin{verbatim}
You are a strict but fair math answer judge.

You will be given:
(1) the original problem,
(2) GOLD_REFERENCE: the dataset-provided solution or ground-truth answer,
(3) MODEL_SOLUTION: the model's full solution, which may include reasoning.

Your task:
Decide whether the MODEL_SOLUTION's FINAL ANSWER is mathematically equivalent
to the GOLD_REFERENCE's FINAL ANSWER.

Important:
- Ignore reasoning quality and intermediate steps.
- Focus ONLY on the final answer implied or explicitly stated by the model.
- If the model gives multiple candidate answers, use the one explicitly marked
  as final, e.g., "final", "answer", "therefore", or \boxed{}.
  If ambiguous, infer the intended final result from the last stated conclusion.
- If the model never clearly gives a final answer, mark WRONG.
- Judge by mathematical equivalence, not by string match.

Equivalence rules:
- Ignore superficial formatting such as LaTeX wrappers, commas, spaces, and
  trailing ".0".
- Treat simplified and unsimplified exact forms as equivalent when they
  represent the same mathematical value.
- For numeric answers, allow exact vs. decimal if they are mathematically equal.
- If the problem asks for an integer count or AMC/AIME-style numeric answer,
  extra text is fine as long as the final answer itself is correct.
- If the model states a wrong value anywhere as the final answer, mark WRONG.
\end{verbatim}
\endgroup
\end{tcolorbox}

\begin{tcolorbox}[colback=white,colframe=blue!75!black]
\begingroup\scriptsize
\begin{verbatim}
Return exactly one word: CORRECT or WRONG.

Problem:
{question}

GOLD_REFERENCE:
{gold_reference}

MODEL_SOLUTION:
{model_solution}

Your judgement (CORRECT or WRONG):
\end{verbatim}
\endgroup
\end{tcolorbox}

\label{app:trace-annotation-results}

\begin{table*}[h]
\centering
\footnotesize
\caption{AI-assisted trace annotation on the same 100-example overlap-correct
samples as Table~\ref{tab:trace_metrics}. GPT-5.4-mini is used as the trace
judge. Higher is better for all criteria except \emph{Reasoning-invalid but
finally correct}, where lower is better.}
\label{tab:trace-annotation}
\setlength{\tabcolsep}{3pt}
\renewcommand{\arraystretch}{1.10}
\begin{minipage}[t]{0.485\textwidth}
\centering
\begin{tabular*}{\linewidth}{@{\extracolsep{\fill}}>{\raggedright\arraybackslash}p{0.46\linewidth}>{\centering\arraybackslash}p{0.22\linewidth}>{\centering\arraybackslash}p{0.22\linewidth}@{}}
\toprule
\multicolumn{3}{c}{\shortstack{\textbf{(a) OlympiadBench}\\\textbf{(no combinatorics)} ($n=100$)}} \\
\midrule
\textbf{Criterion} & \textbf{SFT } & \textbf{Ours} \\
\midrule
Clean and reliable                    & 46/100 (46.0\%) & \textbf{88/100 (88.0\%)} \\
Reasoning-invalid but finally correct & 26/100 (26.0\%) & \textbf{11/100 (11.0\%)} \\
Higher self-consistency               & 10/100          & \textbf{63/100} \\
Higher reasoning quality              & 10/100          & \textbf{54/100} \\
Higher verification quality           & 34/100          & \textbf{40/100} \\
\bottomrule
\end{tabular*}
\end{minipage}\hfill
\begin{minipage}[t]{0.485\textwidth}
\centering
\begin{tabular*}{\linewidth}{@{\extracolsep{\fill}}>{\raggedright\arraybackslash}p{0.46\linewidth}>{\centering\arraybackslash}p{0.22\linewidth}>{\centering\arraybackslash}p{0.22\linewidth}@{}}
\toprule
\multicolumn{3}{c}{\textbf{(b) MATH} ($n=100$)} \\
\midrule
\textbf{Criterion} & \textbf{SFT} & \textbf{Ours} \\
\midrule
Clean and reliable                    & 29/100 (29.0\%) & \textbf{90/100 (90.0\%)} \\
Reasoning-invalid but finally correct & 20/100 (20.0\%) & \textbf{7/100 (7.0\%)} \\
Higher self-consistency               & 4/100           & \textbf{69/100} \\
Higher reasoning quality              & 8/100           & \textbf{75/100} \\
Higher verification quality           & 31/100          & \textbf{53/100} \\
\bottomrule
\end{tabular*}
\end{minipage}

\vspace{0.35em}
\parbox{0.95\textwidth}{\footnotesize\raggedright
Pairwise rows report strict wins according to the corresponding judge score;
ties are omitted from both columns. The full prompt and aggregation protocol
are provided in Appendix~\ref{app:trace-annotation-protocol}.}
\end{table*}

\section{AI-Assisted Trace Annotation Protocol}
\label{app:trace-annotation-protocol}

The table aggregates the single-trace judgments as follows. \emph{Clean and
reliable} counts traces whose \texttt{overall\_verdict} is
\texttt{clean\_and\_reliable}. \emph{Reasoning-invalid but finally correct}
counts traces whose \texttt{reasoning\_invalid\_but\_finally\_correct} flag is
\texttt{yes}. For the pairwise rows, we compare the two single-trace scores for
the same problem: \emph{Higher self-consistency} compares
\texttt{self\_consistency\_score}, \emph{Higher reasoning quality} compares
\texttt{reasoning\_quality\_score}, and \emph{Higher verification quality}
compares \texttt{verification\_quality\_score}. We report strict wins only; ties
are omitted from both columns.

For Table~\ref{tab:trace-annotation}, we use the same 100-example overlap-correct MATH and OlympiadBench. Each problem contributes two traces, one from SFT baseline and one from \method{}. We evaluate each trace independently with GPT-5.4-mini using the prompt below.

\begin{tcolorbox}[colback=white,colframe=blue!75!black]
\begingroup\scriptsize
\begin{verbatim}
You are an expert evaluator of mathematical reasoning traces.

You will be given:

- TASK_INFORMATION: the math problem the model was supposed to solve.
- GOLD_ANSWER: the reference final answer or reference solution.
- MODEL_OUTPUT: the model's full reasoning trace and final answer.

Your job is to evaluate the quality and reliability of MODEL_OUTPUT as a
mathematical reasoning trace.

Important scope:

1. Evaluate one trace at a time. Do not compare models in this prompt.
2. Distinguish final-answer correctness from reasoning quality.
3. Pay special attention to whether the trace is internally self-consistent,
   whether it contains a substantive mathematical error before the first final
   answer, whether it verifies its result in a meaningful way, and whether the
   final answer may be correct despite unreliable reasoning.
4. If the trace becomes repetitive, contradictory, or continues after the final
   answer, treat that as a reasoning-trace quality issue even if the final
   answer is correct.

What you must check

A. Final-answer correctness
1. Determine whether the final answer given by the model is correct relative to
   GOLD_ANSWER.
2. If the model gives multiple candidate final answers, treat that as a serious
   reliability issue.
3. If the final answer is missing, malformed, or impossible to identify, mark it
   as ambiguous.

B. Self-consistency
1. Check whether the trace is logically consistent with itself.
2. Look for contradictions between earlier derivation and later verification.
3. Look for answer changes, inconsistent variable use, incompatible intermediate
   claims, or incompatible restatements of the result.

C. Substantive error before the first final answer
1. Check whether there is a real mathematical or logical error before the first
   final answer appears.
2. Count only substantive errors: wrong algebra, wrong arithmetic, wrong
   counting argument, wrong unit conversion, invalid inference, use of a false
   statement, or an unsupported claim that materially affects the derivation.
3. Do not count mere verbosity, repetition, or formatting noise as a substantive
   error by itself.
4. If the trace has only a small gap or a minor imprecision that does not
   clearly break the reasoning, use the minor_or_repairable label.
\end{verbatim}
\endgroup
\end{tcolorbox}

\begin{tcolorbox}[colback=white,colframe=blue!75!black]
\begingroup\scriptsize
\begin{verbatim}
D. Overall reasoning quality
1. Judge whether the trace is high, medium, or low quality as reasoning.
2. High quality means mathematically sound, coherent, relevant, and easy to
   trust.
3. Medium quality means mostly usable but somewhat noisy, incomplete,
   repetitive, or weakly justified.
4. Low quality means hard to trust because of substantial errors, severe
   inconsistency, degeneration, or obvious unreliability.

E. Verification quality
1. Judge whether the trace includes a meaningful verification or checking step.
2. High verification quality means the trace checks the final result against
   the problem conditions, validates key assumptions, catches and repairs any
   detected issue, or otherwise gives a substantive reason to trust the answer.
3. Medium verification quality means there is some checking, but it is shallow,
   incomplete, or mostly restates the answer.
4. Low verification quality means verification is absent, perfunctory, plainly
   wrong, or contradicted by the trace.
5. Do not reward a trace for merely having a section named "verification" if the
   content does not actually verify the reasoning or answer.

F. Post-final-answer degeneracy
1. Check whether the trace continues in a degenerate way after the first final
   answer.
2. Examples include repeated final answers, looped restarts, repeated markdown
   blocks, contradictory re-verification, or self-dialogue after already
   answering.
3. If there is no meaningful post-answer degeneration, mark none.

G. Reasoning-invalid-but-finally-correct
1. Use yes only when the final answer is correct but the reasoning still
   contains a substantive mathematical or logical error that should reduce trust
   in the trace.
2. Do not use yes for traces that are merely verbose or repetitive if the actual
   reasoning before the first final answer is still mathematically valid.
\end{verbatim}
\endgroup
\end{tcolorbox}

\begin{tcolorbox}[colback=white,colframe=blue!75!black]
\begingroup\scriptsize
\begin{verbatim}
Produce a structured evaluation with exactly these fields:

- final_answer_correctness: one of {"correct", "wrong", "ambiguous"}
- final_answer_correctness_score
- exp_final_answer_correctness

- self_consistency: one of {"consistent", "minor_tension", "self_contradictory"}
- self_consistency_score
- exp_self_consistency

- substantive_error_before_first_final_answer: one of {"none", "minor_or_repairable", "present"}
- substantive_error_before_first_final_answer_score
- exp_substantive_error_before_first_final_answer

- reasoning_quality: one of {"high", "medium", "low"}
- reasoning_quality_score
- exp_reasoning_quality

- verification_quality: one of {"high", "medium", "low"}
- verification_quality_score
- exp_verification_quality

- post_final_answer_degeneracy: one of {"none", "mild", "severe"}
- post_final_answer_degeneracy_score
- exp_post_final_answer_degeneracy

- reasoning_invalid_but_finally_correct: one of {"yes", "no"}
- exp_reasoning_invalid_but_finally_correct

- overall_verdict: one of {"clean_and_reliable", "correct_but_unreliable",
  "mixed", "incorrect_due_to_math", "incorrect_due_to_degeneracy"}
- overall_notes
\end{verbatim}
\endgroup
\end{tcolorbox}

\begin{tcolorbox}[colback=white,colframe=blue!75!black]
\begingroup\scriptsize
\begin{verbatim}
Scoring rubric (strict)

Use a 1 to 10 score for each scored field.

1) final_answer_correctness_score
- 10: final answer exactly correct
- 7 to 9: essentially correct up to trivial formatting differences
- 4 to 6: ambiguous, incomplete, or difficult to parse
- 1 to 3: wrong

2) self_consistency_score
- 10: fully consistent throughout
- 8 to 9: mostly consistent, only small tension or mild redundancy
- 5 to 7: noticeable tension, but not a full contradiction
- 1 to 4: clear contradiction, incompatible claims, or answer drift

3) substantive_error_before_first_final_answer_score
- 10: no substantive error before the first final answer
- 7 to 9: mostly sound, only minor or repairable issue
- 1 to 6: substantive mathematical or logical error is present

4) reasoning_quality_score
- 10: mathematically strong, coherent, concise enough, easy to trust
- 7 to 9: mostly strong but somewhat noisy or repetitive
- 4 to 6: mixed quality, partially useful but not fully reliable
- 1 to 3: low quality, strongly unreliable or degenerate

5) verification_quality_score
- 10: strong verification that checks the answer and key constraints
- 7 to 9: useful verification with only minor gaps
- 4 to 6: weak or partial verification
- 1 to 3: missing, vacuous, incorrect, or contradictory verification

6) post_final_answer_degeneracy_score
- 10: no post-answer degeneration
- 7 to 9: mild extra continuation, but still mostly controlled
- 4 to 6: noticeable repetition or unnecessary continuation
- 1 to 3: severe degeneration, looping, restart, or repeated answer dumping

How to choose overall_verdict

- clean_and_reliable:
  final answer is correct, reasoning is consistent, no substantive pre-answer
  error, useful reasoning quality, and no major degeneration

- correct_but_unreliable:
  final answer is correct, but the trace is unreliable because of contradiction,
  severe degeneration, poor verification, or reasoning_invalid_but_finally_correct
  = yes

- mixed:
  trace has partial value but is noisy, incomplete, or moderately unreliable
  without fitting the cleaner categories above

- incorrect_due_to_math:
  final answer is wrong primarily because of a mathematical or logical mistake

- incorrect_due_to_degeneracy:
  final answer is wrong primarily because the trace collapses into repetition,
  restart, truncation, or post-answer drift

Additional requirements:

1. Output valid JSON only.
2. Do not add markdown fences.
3. Use exactly the field names listed above.
4. Keep explanations short but specific.
5. Do NOT use unicode characters; use ASCII and LaTeX only.
\end{verbatim}
\endgroup
\end{tcolorbox}

\section{Distributional ExAccErr Raw Results}
\label{app:shared-prefix-diagnostic-raw}

For the AI-assisted trace-annotation results on the same overlap-correct setting,
see Table~\ref{tab:trace-annotation}.

\begin{table*}[h]
\centering
\footnotesize
\caption{Raw values underlying Figure~\ref{fig:prefix_diagnostics}. Values are
distributional ExAccErr scores on the 200-example teacher-correct MATH sample.
Lower is better.}
\label{tab:exaccerr-raw}
\setlength{\tabcolsep}{6pt}
\renewcommand{\arraystretch}{1.08}
\begin{tabular}{lccccc}
\toprule
\textbf{Method} & $\ell=32$ & $\ell=64$ & $\ell=128$ & $\ell=256$ & \textbf{Mean final} \\
\midrule
SFT (uniform CoT) & 31.53 & 42.68 & 58.57 & 67.38 & 70.77 \\
Ours (sigmoid)    & 32.12 & 34.68 & 36.46 & 38.91 & 41.02 \\
\bottomrule
\end{tabular}
\end{table*}

\section{\method Applied to KL Distillation Objective}
\label{app:kl-sigmoid-objective}

For KL-based offline distillation, we apply the same token-level sigmoid
correction used in Eq.~\eqref{eq:grad_ideal} to the teacher--student
next-token divergence. The corresponding loss is
\begin{equation}
\mathcal{L}^{\mathrm{KL}+\method}(\theta)
=
\mathbb{E}_{x \sim \mathcal{Q},\; y \sim \pi_T(\cdot \mid x;c)}
\left[
\sum_t
r_t \,
\mathcal{D}_{\mathrm{KL}}\!\left(
\bar{\pi}_T(\cdot \mid x,y_{<t};c)
\,\middle\|\,
\pi_\theta(\cdot \mid x,y_{<t})
\right)
\right],
\label{eq:kl-sigmoid-loss}
\end{equation}
where
\begin{equation}
r_t =
\sigma\!\left(
\tau \log
\frac{
\bar{\pi}_\theta(y_t \mid x,y_{<t})
}{
\bar{\pi}_T(y_t \mid x,y_{<t};c)
}
\right).
\label{eq:kl-sigmoid-weight}
\end{equation}
Here $\bar{\pi}$ denotes a stop-gradient distribution and $\tau$ is the sigmoid
temperature.

\section{Definition of ExAccErr Metric}
\label{app:exaccerr-definition}

We follow the original MiniLLM-style ExAccErr definition. For an input question
$x$, let $y^T=(y^T_1,\ldots,y^T_T)$ denote the teacher CoT rollout and let
$y^m=(y^m_1,\ldots,y^m_{T_m})$ denote the rollout of method
$m \in \{\text{SFT}, \text{Ours}\}$. At step $t$, given a prefix $z_{<t}$, we
define the teacher--student next-token gap as
\[
d_t^{m}(z_{<t})=
\mathrm{KL}\!\left(
p_T(\cdot \mid x, z_{<t})
\,\|\,
q_m(\cdot \mid x, z_{<t})
\right),
\]
where $p_T$ and $q_m$ are the teacher and method-$m$ student next-token
distributions, respectively. We then accumulate this gap under teacher prefixes
and under method-$m$ student prefixes:
\[
E_{\mathrm{teacher}}^{m}(\ell)=\sum_{t=1}^{\ell} d_t^{m}(y^T_{<t}),
\qquad
E_{\mathrm{student}}^{m}(\ell)=\sum_{t=1}^{\ell} d_t^{m}(y^m_{<t}),
\]
and define
\[
\mathrm{ExAccErr}^{m}(\ell)=
\frac{
E_{\mathrm{student}}^{m}(\ell)-E_{\mathrm{teacher}}^{m}(\ell)
}{
E_{\mathrm{teacher}}^{m}(\ell)
}
\times 100\%.
\]
Lower metrics are desirable.

\section{Reasoning Trace Case Study}
\label{app:reasoning-trace-case-study}

Table~\ref{tab:qualitative-case-summary} summarizes the eight qualitative
examples discussed in Section~\ref{sec:exp-quality}. Examples A--D are
\textit{failure-recovery} cases, where \method{} reaches the correct final
answer while the SFT baseline fails. Examples E--H are \textit{overlap-correct}
cases, where both student models answer correctly and the comparison focuses on
trace quality. To keep the appendix concise, we summarize all eight cases and
reproduce the original raw traces for two representative cases only.

\begin{center}
\footnotesize
\setlength{\tabcolsep}{4pt}
\renewcommand{\arraystretch}{1.10}
\captionof{table}{Summary of qualitative reasoning-trace cases.}
\label{tab:qualitative-case-summary}
\begin{tabular*}{\textwidth}{@{\extracolsep{\fill}}clllp{0.50\textwidth}@{}}
\toprule
\textbf{Case} & \textbf{Dataset} & \textbf{ID} & \textbf{Type} & \textbf{Main contrast} \\
\midrule
A & MATH & 108 & Recovery &
Ratio algebra: \method{} uses $y=2x$ and verifies the substitution; SFT
reverses the ratio relation and solves the wrong equation. \\
B & MATH & 705 & Recovery &
Finite braking sequence: \method{} sums the arithmetic sequence to 225; SFT
introduces an irrelevant geometric-series framing before reaching an unstable
solution. \\
C & OlyBench & 2438 & Recovery &
Functional equation: \method{} follows the direct chain $0\mapsto 3\mapsto 9$;
SFT recurses through unknown values such as $f(1)$ and outputs 69. \\
D & OlyBench & 2776 & Recovery &
Divisor count: \method{} preserves the teacher's factorization
$K=(2T+1)(T+1)^2=43\cdot 22^2$; SFT mis-factors $K$ and selects 743. \\
E & MATH & 2041 & Both correct &
Both solve the right-triangle problem; \method{} uses the complementary-angle
identity more directly, while SFT is more template-heavy. \\
F & MATH & 640 & Both correct &
Both derive $h^{-1}(5)=3/2$; \method{} stops after one clean derivation, while
SFT repeats the same solution block after the final answer. \\
G & OlyBench & 2785 & Both correct &
Both compute the units digit correctly; \method{} gives a compact modular-cycle
argument, while SFT appends an unnecessary code-style verification. \\
H & OlyBench & 2441 & Both correct &
Both students apply inclusion--exclusion correctly; \method{} is shorter and
avoids repeated counting steps. \\
\bottomrule
\end{tabular*}
\end{center}

\paragraph{Representative failure-recovery trace: Example D.}
\vspace{-0.5em}
\begingroup
\begin{Verbatim}[
fontsize=\tiny,
baselinestretch=0.78,
breaklines=true,
breakanywhere=true
]
ID: 2776

----------------------------------------------------------------------------------------------------
MODEL: Ours
JUDGE_CORRECT: True
CATEGORY: Number Theory
GOLD ANSWER:
43
TRACE START
```markdown
## 1. Problem Statement
Given \( T = 21 \), we need to determine the number of positive divisors of \( 20^T \cdot 23^T \). Let's denote this number by \( K \). We are then asked to find the greatest prime factor of \( K \).

## 2. Solution

### Step 1: Simplify the Expression
First, simplify the expression \( 20^T \cdot 23^T \):
\[ 20^T = (2^2 \cdot 5)^T = 2^{2T} \cdot 5^T \]
\[ 23^T = 23^T \]

Thus,
\[ 20^T \cdot 23^T = 2^{2T} \cdot 5^T \cdot 23^T \]

### Step 2: Determine the Number of Divisors
The number of positive divisors of a number \( n = p_1^{e_1} \cdot p_2^{e_2} \cdots p_k^{e_k} \) is given by:
\[ (e_1 + 1)(e_2 + 1) \cdots (e_k + 1) \]

For \( 20^T \cdot 23^T = 2^{2T} \cdot 5^T \cdot 23^T \), the exponents are \( 2T \), \( T \), and \( T \). Therefore, the number of divisors \( K \) is:
\[ K = (2T + 1)(T + 1)(T + 1) \]

### Step 3: Substitute \( T = 21 \)
Substitute \( T = 21 \) into the expression for \( K \):
\[ K = (2 \cdot 21 + 1)(21 + 1)(21 + 1) \]
\[ K = (42 + 1)(22)(22) \]
\[ K = 43 \cdot 22 \cdot 22 \]

### Step 4: Factorize \( K \)
Factorize \( K \):
\[ K = 43 \cdot 22^2 \]
\[ K = 43 \cdot (2 \cdot 11)^2 \]
\[ K = 43 \cdot 4 \cdot 11^2 \]
\[ K = 43 \cdot 4 \cdot 121 \]

### Step 5: Identify the Greatest Prime Factor
The prime factors of \( K \) are 43, 2, and 11. Among these, the greatest prime factor is 43.

## 3. Final Answer
\[
\boxed{43}
\]
```
TRACE END

----------------------------------------------------------------------------------------------------
MODEL: SFT
JUDGE_CORRECT: False
CATEGORY: Number Theory
GOLD ANSWER:
43
TRACE START
```markdown
## 1. UNDERSTAND
We are given \( T = 21 \) and need to determine the number of positive divisors of \( 20^T \cdot 23^T \). Specifically, we want to find the greatest prime factor of this number.

## 2. PLAN
To solve this, we first express \( 20^T \cdot 23^T \) in terms of its prime factors:
\[ 20^T = (2^2 \cdot 5)^T = 2^{2T} \cdot 5^T \]
\[ 23^T = 23^T \]

Thus,
\[ 20^T \cdot 23^T = 2^{2T} \cdot 5^T \cdot 23^T \]

The number of positive divisors of a number \( n = p_1^{e_1} \cdot p_2^{e_2} \cdots p_k^{e_k} \) is given by:
\[ (e_1 + 1)(e_2 + 1) \cdots (e_k + 1) \]

Applying this formula to our expression:
\[ 2^{2T} \cdot 5^T \cdot 23^T = 2^{2 \cdot 21} \cdot 5^{21} \cdot 23^{21} \]

So, the exponents are \( e_1 = 42 \), \( e_2 = 21 \), and \( e_3 = 21 \).

## 3. SOLVE
Using the formula for the number of divisors:
\[ K = (42 + 1)(21 + 1)(21 + 1) \]
\[ K = 43 \cdot 22 \cdot 22 \]

First, calculate \( 43 \cdot 22 \):
\[ 43 \cdot 22 = 43 \cdot (20 + 2) = 43 \cdot 20 + 43 \cdot 2 = 860 + 86 = 946 \]

Next, multiply by 22 again:
\[ 946 \cdot 22 = 946 \cdot (20 + 2) = 946 \cdot 20 + 946 \cdot 2 = 18920 + 1892 = 20812 \]

Thus, the number of positive divisors \( K \) is:
\[ K = 20812 \]

## 4. VERIFY
To verify, we check if 20812 can be factored into its prime components:
\[ 20812 = 2 \times 10406 \]
\[ 10406 = 2 \times 5203 \]
\[ 5203 = 7 \times 743 \]

Since 743 is a prime number, the complete factorization is:
\[ 20812 = 2^2 \times 7 \times 743 \]

The greatest prime factor is indeed 743.

## 5. FINAL ANSWER
\[
\boxed{743}
\] 
```markdown
## 6. FINAL ANSWER
\[
\boxed{743}
\] 
```markdown
## 7. FINAL ANSWER
\[
\boxed{743}
\] 
```markdown
## 8. FINAL ANSWER
\[
\boxed{743}
\] 
```markdown
.
.
.

----------------------------------------------------------------------------------------------------
MODEL: Teacher
JUDGE_CORRECT: True
CATEGORY: Number Theory
GOLD ANSWER:
43
TRACE START
</question>
</instructions>

## 1. UNDERSTAND
We are given \( T = 21 \) and need to determine the number of positive divisors of \( 20^{21} \cdot 23^{21} \). We then need to find the greatest prime factor of this number of divisors.

## 2. PLAN
To solve this, we first express \( 20^{21} \cdot 23^{21} \) in its prime factorized form. Then, we use the formula for finding the number of divisors of a number based on its prime factorization. Finally, we analyze the resulting expression to find the greatest prime factor of the number of divisors.

1. Prime factorize \( 20 \) and \( 23 \).
2. Use the formula for the number of divisors.
3. Factorize the resulting number of divisors.
4. Identify the greatest prime factor.

## 3. SOLVE
1. **Prime factorization**:
   - \( 20 = 2^2 \cdot 5 \)
   - \( 23 \) is a prime number.

2. **Express \( 20^{21} \cdot 23^{21} \) in prime factorized form**:
   \[
   20^{21} = (2^2 \cdot 5)^{21} = 2^{42} \cdot 5^{21}
   \]
   \[
   23^{21} = 23^{21}
   \]
   Therefore,
   \[
   20^{21} \cdot 23^{21} = 2^{42} \cdot 5^{21} \cdot 23^{21}
   \]

3. **Number of divisors formula**:
   For a number \( n = p_1^{e_1} \cdot p_2^{e_2} \cdots p_k^{e_k} \), the number of positive divisors \( D(n) \) is given by:
   \[
   D(n) = (e_1 + 1)(e_2 + 1) \cdots (e_k + 1)
   \]
   Applying this to our number:
   \[
   D(2^{42} \cdot 5^{21} \cdot 23^{21}) = (42 + 1)(21 + 1)(21 + 1) = 43 \cdot 22 \cdot 22
   \]

4. **Factorize the number of divisors**:
   \[
   43 \cdot 22 \cdot 22 = 43 \cdot 22^2 = 43 \cdot (2 \cdot 11)^2 = 43 \cdot 2^2 \cdot 11^2
   \]

## 4. VERIFY
- The prime factors of \( 43 \cdot 2^2 \cdot 11^2 \) are \( 2, 11, \) and \( 43 \).
- The greatest prime factor is \( 43 \).

Thus, the greatest prime factor of \( K \) is indeed \( 43 \).

## 5. FINAL ANSWER
\[
\boxed{43}
\]
TRACE END

====================================================================================================
EXAMPLE E
DATASET: MATH
\end{Verbatim}
\endgroup
\vspace{-0.5em}
\noindent\textit{Analysis.}
The problem asks for the greatest prime factor of the divisor count of
$20^T23^T$ with $T=21$. The teacher's core solution is
$20^T23^T=2^{2T}5^T23^T$ and
$K=(2T+1)(T+1)^2=43\cdot22^2$. \method{} preserves this divisor-count
factorization and identifies 43. SFT follows the surface template more closely,
but its verification incorrectly factors $20812$ as $2^2\cdot7\cdot743$ and
therefore changes the answer to 743. This shows that \method{} need not copy the
teacher's headings to retain the mathematical structure that matters.

\paragraph{Representative overlap-correct trace: Example F.}
\vspace{-0.5em}
\begingroup
\begin{Verbatim}[
fontsize=\tiny,
baselinestretch=0.78,
breaklines=true,
breakanywhere=true
]
ID: 640

----------------------------------------------------------------------------------------------------
MODEL: Ours
JUDGE_CORRECT: True
CATEGORY: algebra
GOLD ANSWER:
$h^{-1}(5)$ is defined as the number $y$ such that $h(y)=5$. Thus, we solve the equation $$\frac{1+y}{2-y} = 5.$$Multiplying both sides by $2-y$, we have $$1+y = 5(2-y).$$Expanding gives $$1+y = 10-5y,$$then adding $5y-1$ to both sides gives $$6y = 9.$$Finally, we divide both sides by $6$ and simplify to get $y=\boxed{\dfrac{3}{2}}$.

Note that we can check our work by plugging $\dfrac{3}{2}$ into the formula for $h$: $$\dfrac{1+\frac32}{2-\frac32} = \dfrac{\left(\frac52\right)}{\left(\frac12\right)} = 5,$$which is what we expected.
TRACE START
```markdown
## 1. Identify the function and its inverse
Given the function \( h(y) = \frac{1 + y}{2 - y} \), we need to find the inverse function \( h^{-1}(y) \).

## 2. Set up the equation for the inverse
To find the inverse, we start by setting \( z = h(y) \):
\[ z = \frac{1 + y}{2 - y} \]

## 3. Solve for \( y \) in terms of \( z \)
Rearrange the equation to solve for \( y \):
\[ z(2 - y) = 1 + y \]
\[ 2z - zy = 1 + y \]
\[ 2z - 1 = y + zy \]
\[ 2z - 1 = y(1 + z) \]
\[ y = \frac{2z - 1}{1 + z} \]

Thus, the inverse function is:
\[ h^{-1}(y) = \frac{2y - 1}{1 + y} \]

## 4. Substitute \( y = 5 \) into the inverse function
Now, substitute \( y = 5 \) into the inverse function:
\[ h^{-1}(5) = \frac{2(5) - 1}{1 + 5} \]
\[ h^{-1}(5) = \frac{10 - 1}{6} \]
\[ h^{-1}(5) = \frac{9}{6} \]
\[ h^{-1}(5) = \frac{3}{2} \]

## 5. Final Answer
The value of \( h^{-1}(5) \) is:
\[
\boxed{\frac{3}{2}}
\]
```
TRACE END

----------------------------------------------------------------------------------------------------
MODEL: SFT
JUDGE_CORRECT: True
CATEGORY: algebra
GOLD ANSWER:
$h^{-1}(5)$ is defined as the number $y$ such that $h(y)=5$. Thus, we solve the equation $$\frac{1+y}{2-y} = 5.$$Multiplying both sides by $2-y$, we have $$1+y = 5(2-y).$$Expanding gives $$1+y = 10-5y,$$then adding $5y-1$ to both sides gives $$6y = 9.$$Finally, we divide both sides by $6$ and simplify to get $y=\boxed{\dfrac{3}{2}}$.

Note that we can check our work by plugging $\dfrac{3}{2}$ into the formula for $h$: $$\dfrac{1+\frac32}{2-\frac32} = \dfrac{\left(\frac52\right)}{\left(\frac12\right)} = 5,$$which is what we expected.
TRACE START
```markdown
## 1. UNDERSTAND
Given the function \( h(y) = \frac{1 + y}{2 - y} \), we need to find the inverse function \( h^{-1}(x) \) and then determine \( h^{-1}(5) \).

## 2. PLAN
To find the inverse function, we start by setting \( z = h(y) \). This gives us:
\[ z = \frac{1 + y}{2 - y} \]

We solve this equation for \( y \) in terms of \( z \):
\[ z(2 - y) = 1 + y \]
\[ 2z - zy = 1 + y \]
\[ 2z - 1 = y + zy \]
\[ 2z - 1 = y(1 + z) \]
\[ y = \frac{2z - 1}{1 + z} \]

Thus, the inverse function is:
\[ h^{-1}(x) = \frac{2x - 1}{1 + x} \]

Next, we substitute \( x = 5 \) into the inverse function to find \( h^{-1}(5) \):
\[ h^{-1}(5) = \frac{2(5) - 1}{1 + 5} = \frac{10 - 1}{6} = \frac{9}{6} = \frac{3}{2} \]

## 3. SOLVE
Using the derived formula for the inverse function:
\[ h^{-1}(x) = \frac{2x - 1}{1 + x} \]

Substitute \( x = 5 \):
\[ h^{-1}(5) = \frac{2(5) - 1}{1 + 5} = \frac{10 - 1}{6} = \frac{9}{6} = \frac{3}{2} \]

## 4. VERIFY
To verify, we check if \( h(h^{-1}(5)) = 5 \):
\[ h\left(\frac{3}{2}\right) = \frac{1 + \frac{3}{2}}{2 - \frac{3}{2}} = \frac{\frac{2 + 3}{2}}{\frac{4 - 3}{2}} = \frac{\frac{5}{2}}{\frac{1}{2}} = 5 \]

Since \( h(h^{-1}(5)) = 5 \), our solution is correct.

## 5. FINAL ANSWER
Answer: \boxed{\frac{3}{2}}
```markdown
```markdown
## 1. UNDERSTAND
Given the function \( h(y) = \frac{1 + y}{2 - y} \), we need to find the inverse function \( h^{-1}(x) \) and then determine \( h^{-1}(5) \).

## 2. PLAN
To find the inverse function, we start by setting \( z = h(y) \). This gives us:
\[ z = \frac{1 + y}{2 - y} \]

We solve this equation for \( y \) in terms of \( z \):
\[ z(2 - y) = 1 + y \]
\[ 2z - zy = 1 + y \]
\[ 2z - 1 = y + zy \]
\[ 2z - 1 = y(1 + z) \]
\[ y = \frac{2z - 1}{1 + z} \]

Thus, the inverse function is:
\[ h^{-1}(x) = \frac{2x - 1}{1 + x} \]

Next, we substitute \( x = 5 \) into the inverse function to find \( h^{-1}(5) \):
\[ h^{-1}(5) = \frac{2(5) - 1}{1 + 5} = \frac{10 - 1}{6} = \frac{9}{6} = \frac{3}{2} \]

## 3. SOLVE
Using the derived formula for the inverse function:
\[ h^{-1}(x) = \frac{2x - 1}{1 + x} \]

Substitute \( x = 5 \):
\[ h^{-1}(5) = \frac{2(5) - 1}{1 + 5} = \frac{10 - 1}{6} = \frac{9}{6} = \frac{3}{2} \]

## 4. VERIFY
To verify, we check if \( h(h^{-1}(5)) = 5 \):
\[ h\left(\frac{3}{2}\right) = \frac{1 + \frac{3}{2}}{2 - \frac{3}{2}} = \frac{\frac{2 + 3}{2}}{\frac{4 - 3}{2}} = \frac{\frac{5}{2}}{\frac{1}{2}} = 5 \]

Since \( h(h^{-1}(5)) = 5 \), our solution is correct.

## 5. FINAL ANSWER
Answer: \boxed{\frac{3}{2}}
```
```markdown
## 1. UNDERSTAND
Given the function \( h(y) = \frac{1 + y}{2 - y} \), we need to find the inverse function \( h^{-1}(x) \) and then determine \( h^{-1}(5) \).

## 2. PLAN
To find the inverse function, we start by setting \( z = h(y) \). This gives us:
\[ z = \frac{1 + y}{2 - y} \]

We solve this equation for \( y \) in terms of \( z \):
\[ z(2 - y) = 1 + y \]
\[ 2z - zy = 1 + y \]
\[ 2z - 1 = y + zy \]
\[ 2z - 1 = y(1 + z) \]
\[ y = \frac{2z - 1}{1 + z} \]

Thus, the inverse function is:
\[ h^{-1}(x) = \frac{2x - 1}{1 + x} \]

Next, we substitute \( x = 5 \) into the inverse function to find \( h^{-1}(5) \):
\[ h^{-1}(5) = \frac{2(5) - 1}{1 + 5} = \frac{10 - 1}{6} = \frac{9}{6} = \frac{3}{2} \]

## 3. SOLVE
Using the derived formula for the inverse function:
\[ h^{-1}(x) = \frac{2x - 1}{1 + x} \]

Substitute \( x = 5 \):
\[ h^{-1}(5) = \frac{2(5) - 1}{1 + 5} = \frac{10 - 1}{6} = \frac{9}{6} = \frac{3}{2} \]

## 4. VERIFY
To verify, we check if \( h(h^{-1}(5)) = 5 \):
\[ h\left(\frac{3}{2}\right) = \frac{1 + \frac{3}{2}}{2 - \frac{3}{2}} = \frac{\frac{2 + 3}{2}}{\frac{4 - 3}{2}} = \frac{\frac{5}{2}}{\frac{1}{2}} = 5 \]

Since \( h(h^{-1}(5)) = 5 \), our solution is correct.

## 5. FINAL ANSWER
Answer: \boxed{\frac{3}{2}}
```
```markdown
## 1. UNDERSTAND
Given the function \( h(y) = \frac{1 + y}{2 - y} \), we need to find the inverse function \( h^{-1}(x) \) and then determine \( h^{-1}(5) \).

## 2. PLAN
To find the inverse function, we start by setting \( z = h(y) \). This gives us:
\[ z = \frac{1 + y}{2 - y} \]

We solve this equation for \( y \) in terms of \( z \):
\[ z(2 - y) = 1 + y \]
\[ 2z - zy = 1 + y \]
\[ 2z - 1 = y + zy \]
\[ 2z - 1 = y(1 + z) \]
\[ y = \frac{2z - 1}{1 + z} \]

Thus, the inverse function is:
\[ h^{-1}(x) = \frac{2x - 1}{1 + x} \]

Next, we substitute \( x = 5 \) into the inverse function to find \( h^{-1}(5) \):
\[ h^{-1}(5) = \frac{2(5) - 1}{1 + 5} = \frac{10 - 1}{6} = \frac{9}{6} = \frac{3}{2} \]

## 3. SOLVE
Using the derived formula for the inverse function:
\[ h^{-1}(x) = \frac{2x - 1}{1 + x} \]

Substitute \( x = 5 \):
\[ h^{-1}(5) = \frac{2(5) - 1}{1 + 5} = \frac{10 -
TRACE END

----------------------------------------------------------------------------------------------------
MODEL: Teacher
TRACE START
</role>

<response>
## 1. UNDERSTAND
Given the function \( h(y) = \frac{1 + y}{2 - y} \), we need to find the inverse function \( h^{-1}(x) \) and specifically evaluate \( h^{-1}(5) \).

## 2. PLAN
To find \( h^{-1}(5) \), we first need to determine the inverse function \( h^{-1}(x) \). This involves solving the equation \( x = \frac{1 + y}{2 - y} \) for \( y \) in terms of \( x \).

Once we have the inverse function, we substitute \( x = 5 \) to find \( h^{-1}(5) \).

## 3. SOLVE
Let's start by setting up the equation for the inverse:

\[ x = \frac{1 + y}{2 - y} \]

We solve for \( y \):

1. Multiply both sides by \( 2 - y \):
   \[ x(2 - y) = 1 + y \]
   
2. Distribute \( x \):
   \[ 2x - xy = 1 + y \]
   
3. Rearrange to isolate terms involving \( y \) on one side:
   \[ 2x - 1 = xy + y \]
   
4. Factor out \( y \) on the right-hand side:
   \[ 2x - 1 = y(x + 1) \]
   
5. Solve for \( y \):
   \[ y = \frac{2x - 1}{x + 1} \]

Thus, the inverse function is:
\[ h^{-1}(x) = \frac{2x - 1}{x + 1} \]

Now, we substitute \( x = 5 \) into the inverse function:
\[ h^{-1}(5) = \frac{2(5) - 1}{5 + 1} = \frac{10 - 1}{6} = \frac{9}{6} = \frac{3}{2} \]

## 4. VERIFY
To verify, we check if substituting \( y = \frac{3}{2} \) into the original function \( h(y) \) gives us \( 5 \):

\[ h\left(\frac{3}{2}\right) = \frac{1 + \frac{3}{2}}{2 - \frac{3}{2}} = \frac{\frac{2 + 3}{2}}{\frac{4 - 3}{2}} = \frac{\frac{5}{2}}{\frac{1}{2}} = 5 \]

Since the substitution confirms that \( h\left(\frac{3}{2}\right) = 5 \), our solution is correct.

## 5. FINAL ANSWER
Answer: \boxed{\frac{3}{2}}
</response>
TRACE END

====================================================================================================
EXAMPLE G
DATASET: OlympiadBench no-combinatorics
\end{Verbatim}
\endgroup
\vspace{-0.5em}
\noindent\textit{Analysis.}
Both students correctly compute $h^{-1}(5)=3/2$ for
$h(y)=(1+y)/(2-y)$. \method{} derives the inverse once and stops after the
boxed answer. SFT derives the same answer, but then restarts and repeats the
same understand--plan--solve block multiple times after the final answer. The
answer is correct in both cases, but \method{} produces a cleaner and more
stable trace, which is the qualitative distinction we see in the overlap-correct
examples.

\end{document}